\documentclass[journal]{IEEEtran}
\usepackage{amsmath,amsfonts,graphicx,epsfig}
\usepackage{algorithmic}
\usepackage{algorithm}
\usepackage{array}
\usepackage[caption=false,font=normalsize,labelfont=sf,textfont=sf]{subfig}
\usepackage{textcomp}
\usepackage{stfloats}
\usepackage{verbatim}
\usepackage{arydshln}
\usepackage{multirow}
\usepackage{booktabs}
\usepackage{float}
\usepackage{makecell}
\usepackage{longtable}
\usepackage{threeparttable}
\usepackage{diagbox}
\usepackage{url}
\usepackage{amssymb}
\usepackage{enumitem}
\usepackage{graphicx}
\usepackage{hyperref}
\usepackage{xcolor}
\usepackage{color}
\usepackage{marvosym}

\usepackage{fontawesome}

\hypersetup{
    colorlinks=true,
    linkcolor=black,
    filecolor=black,      
    urlcolor=black,
    citecolor=black,
}
\usepackage{cite}
\usepackage{booktabs} 
\newcommand{\ieno}{\textit{i.e.}}
\newcommand{\egno}{\textit{e.g.}}
\newsavebox\myboxA
\newsavebox\myboxB
\newlength\mylenA
\newcommand*\xoverline[2][0.75]{%
\sbox{\myboxA}{$\m@th#2$}%
\setbox\myboxB\null% Phantom box
\ht\myboxB=\ht\myboxA%
\dp\myboxB=\dp\myboxA%
\wd\myboxB=#1\wd\myboxA% Scale phantom
\sbox\myboxB{$\m@th\overline{\copy\myboxB}$}% Overlined phantom
\setlength\mylenA{\the\wd\myboxA}% calc width diff
\addtolength\mylenA{-\the\wd\myboxB}%
\ifdim\wd\myboxB<\wd\myboxA%
\rlap{\hskip 0.5\mylenA\usebox\myboxB}{\usebox\myboxA}%
\else
\hskip -0.5\mylenA\rlap{\usebox\myboxA}{\hskip 0.5\mylenA\usebox\myboxB}%
\fi}
\usepackage{soul}
\usepackage{xcolor}
\setstcolor{red}
\soulregister\cite7
\soulregister\ref7 

\begin{document}

\title{A Survey on All-in-One Image Restoration: Taxonomy, Evaluation and Future Trends}

\author{Junjun~Jiang\textsuperscript{\Letter},~\IEEEmembership{Senior Member,~IEEE},~%\textsuperscript{\Letter}
        Zengyuan Zuo,~
        Gang~Wu,~
        Kui Jiang,~
        and Xianming Liu,~\IEEEmembership{Member,~IEEE}

\IEEEcompsocitemizethanks{

\IEEEcompsocthanksitem J. Jiang, Z. Zuo, G. Wu, K. Jiang, and X. Liu are with the School of Computer Science and Technology, Harbin Institute of Technology, Harbin 150001, China. E-mail: \{jiangjunjun@hit.edu.cn, 24s103286@stu.hit.edu.cn, gwu@hit.edu.cn, jiangkui@hit.edu.cn, csxm@hit.edu.cn\}. Corresponding author: Junjun Jiang. 
% \IEEEcompsocthanksitem Copyright (c) 2013 IEEE. Personal use of this material is permitted. However, permission to use this material for any other purposes must be obtained from the IEEE by sending a request to pubs-permissions@ieee.org. \protect\\
}
\thanks{The research was supported by the National Natural Science Foundation of China (U23B2009, 62471158).} 

}

\markboth{Journal of \LaTeX\ Class Files,~Vol.~14, No.~8, August~2021}%
{Shell \MakeLowercase{\textit{et al.}}: A Comprehensive Survey of All-in-One Image Restoration}

\maketitle

\begin{abstract}
Image restoration (IR) seeks to recover high-quality images from degraded observations caused by a wide range of factors, including noise, blur, compression, and adverse weather. While traditional IR methods have made notable progress by targeting individual degradation types, their specialization often comes at the cost of generalization, leaving them ill-equipped to handle the multifaceted distortions encountered in real-world applications. In response to this challenge, the all-in-one image restoration (AiOIR) paradigm has recently emerged, offering a unified framework that adeptly addresses multiple degradation types. These innovative models enhance the convenience and versatility by adaptively learning degradation-specific features while simultaneously leveraging shared knowledge across diverse corruptions. In this survey, we provide the first in-depth and systematic overview of AiOIR, delivering a structured taxonomy that categorizes existing methods by architectural designs, learning paradigms, and their core innovations. We systematically categorize current approaches and assess the challenges these models encounter, outlining research directions to propel this rapidly evolving field. To facilitate the evaluation of existing methods, we also consolidate widely-used datasets, evaluation protocols, and implementation practices, and compare and summarize the most advanced open-source models. As the first comprehensive review dedicated to AiOIR, this paper aims to map the conceptual landscape, synthesize prevailing techniques, and ignite further exploration toward more intelligent, unified, and adaptable visual restoration systems. A curated code repository is available at \url{https://github.com/Harbinzzy/All-in-One-Image-Restoration-Survey}.
\end{abstract}


% \section{Introduction}
% \label{sec:introduction}
\begin{figure*}[ht]
	\centering 
	\includegraphics[width=\linewidth]{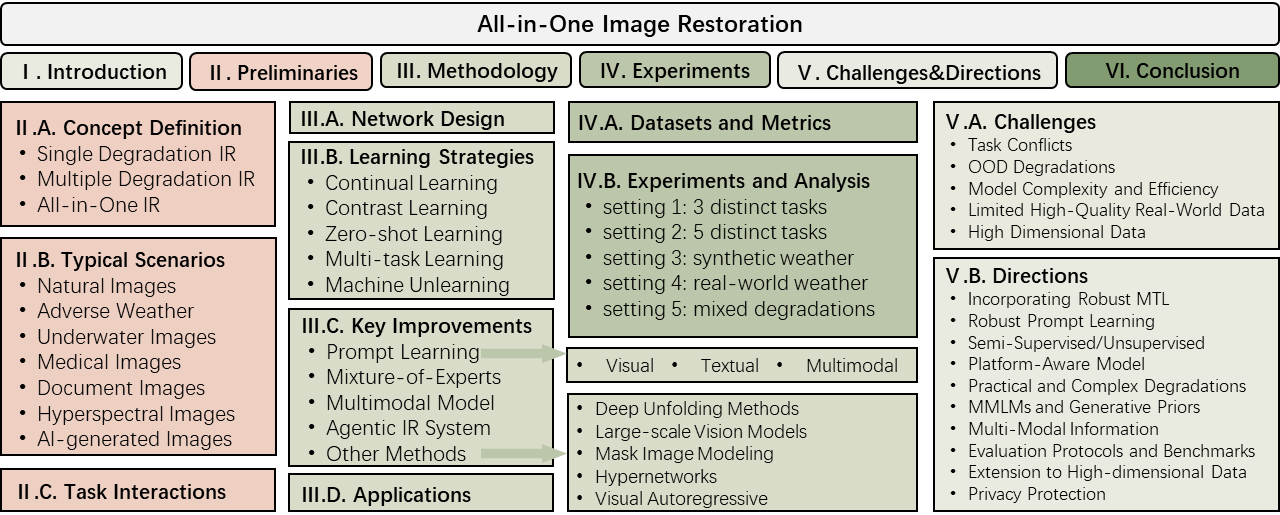}
	\caption{Hierarchically-structured taxonomy of this survey.}
	\label{fig:all}
\end{figure*}
\section{Introduction}

\IEEEPARstart{I}{mage} processing is an integral part of low-level vision tasks, and digital image processing has evolved significantly over the past few decades, transitioning from traditional methods to advanced deep learning techniques. %Initially, image processing relied heavily on algorithms that performed tasks such as filtering, edge detection, image synthesis and image segmentation. 
These methods, while effective, are limited by their inability to handle complex and varied image degradation scenarios. With the rise of deep learning, image processing has achieved remarkable results, especially driven by convolutional neural networks (CNNs)~\cite{krizhevsky2017imagenet-cnn1} and Transformers~\cite{vaswani2017attention-transformer}. In the image restoration (IR) field, single-task IR has achieved notable breakthroughs in correcting a specific type of image degradation (\egno, denoising~\cite{dabov2007color,liu2020lira,tian2020image,zhang2017beyond,zhang2017learning,zhang2018ffdnet}, dehazing~\cite{cai2016dehazenet,dong2020fd,fan2016two,he2010single,li2021you}, desnowing~\cite{chen2021all,chen2023uncertainty, wang2023smartassign}, deraining~\cite{ren2019progressive, yang2020single,DRSformer}, deblurring~\cite{tao2018scale,yuan2007image} and low-light image enhancement~\cite{guo2016lime,ma2022toward,wang2022lowl}). Although they have set state-of-the-art performance in restoring diverse degraded images, existing single-task IR methods generally lack the flexibility to adapt to new types of image degradation without extensive retraining (it is worth noting that some real-world IR methods~\cite{wang2021real, yue2023resshift, wang2024exploiting, yu2024scaling, zhou2025unires} are not categorized as single-task IR). 
To alleviate the aforementioned issues, researchers have proposed all-in-one image restoration (AiOIR), devoting significant effort to introduce various key improvements (\egno, prompt-based learning approach, mixture-of-experts structure, multimodal model and autonomous agent) beyond single-task IR.

The AiOIR approach aims to develop a model capable of addressing multiple isolated or mixed degradation tasks (e.g., denoising, deblurring, dehazing, deraining, and low-light enhancement) within a unified framework without the need for task-specific retraining. While single-task models often achieve state-of-the-art performance for specific restoration tasks, AiOIR models offer the potential for more efficient deployment and unified processing, particularly when handling unknown, mixed, or dynamically changing degradations. Despite the variety of AiOIR models, their ability to generate high-quality images is still under active exploration. Recently, pioneering researchers have been investigating ways to optimize the architectures of these models to balance the trade-offs between computational complexity and restoration quality. 

With the rapid development of AiOIR, numerous researchers have collected a series of datasets tailored for various all-in-one task settings, \egno, the All-Weather dataset~\cite{valanarasu2022transweather} for synthetic weather-related scenarios, the WeatherStream dataset~\cite{Zhang2023WeatherStreamLT} for real-world weather-related scenarios, the CDD-11 dataset~\cite{guo2024onerestore} for composite degradations, \emph{etc}. Leveraging these datasets, most recent works focus on improving the representation capability of IR networks for complicated degradation through well-designed methods based on prompt learning, contrastive learning, multimodal representations, \emph{etc}. Although these works have achieved superior progress in the objective quality (\egno, PSNR, NIQE~\cite{IL-NIQE}, and FID \cite{sheikh2005information}), the restored images still suffer from unsatisfactory texture generation, hindering the practical deployment of IR methods in real-world scenarios. Collectively, these methodologies represent a substantial progression in the pursuit of sophisticated, accurate, and versatile AiOIR solutions.

AiOIR methods have emerged as a positive step forward, although their specific implementations and applications are still under exploration. While there are comprehensive surveys for single-task IR methods, such as image super-resolution~\cite{9044873,anwar2020deep,chen2022real,liu2022blind}, deraining~\cite{wang2022survey,su2023survey}, dehazing~\cite{10080156}, denoising~\cite{tian2020deep,goyal2020image,elad2023image}, deblurring~\cite{7479956, 10074406}, low-light enhancement~\cite{liu2021benchmarking,li2021low,wang2022low}, and the review covering diffusion-based image restoration~\cite{Li2023DiffusionMF}, no existing review has systematically examined the emerging field of AiOIR. To bridge this gap, we aim to provide the first comprehensive overview of AiOIR methods, shedding light on both its representative approaches and diverse improvements. Fig.~\ref{fig:all} shows the taxonomy of this survey in a hierarchically-structured way. In the following sections, we will cover various aspects of recent advances in AiOIR:
\begin{enumerate}
    \item Preliminaries (Sec.~\ref{sec:Pre}): We introduce the definition of AiOIR and provide a comparison with related concepts. We point out typical scenarios and task relationships.  
    
    \item Methodology (Sec.~\ref{sec:IRmethods}): We provide a detailed analysis of representative AiOIR networks, aiming to illustrate the prevailing methods and elucidate the different categories of methods. By analyzing state-of-the-art methods, we summarize network architectures, learning strategies, some key improvements, and applications.
    
    \item Experiments (Sec.~\ref{sec:exp}): To facilitate a reasonable and exhaustive comparison, we clarify the commonly-used datasets and experimental settings in AiOIR. Furthermore, comprehensive comparisons across different settings are provided.
    
    \item Challenges and Future Directions (Sec.~\ref{sec:future}): There are still some challenges to extending AiOIR to the practical applications. To further improve the development of AiOIR, we summarize the primary challenges and propose the potential directions and trends.
    
\end{enumerate}

\section{Preliminaries}
\label{sec:Pre}
In this section, we first define the concept of AiOIR and compare it with related concepts. Subsequently, we provide a comprehensive review of diverse scenarios in AiOIR and the corresponding approaches. Finally, we illustrate the interactions between different IR tasks.
\subsection{Concept Definitions}
Single-degradation image restoration (SDIR) focuses on recovering clean images from observations that have been corrupted by a specific type of degradation, such as noise~\cite{liu2020lira,tian2020image,zhang2017beyond}, blur~\cite{tao2018scale,yuan2007image}, or rain~\cite{ren2019progressive, yang2020single,DRSformer}. These methods are typically task-specific, meaning that each model is designed for a particular degradation type. While these models demonstrate strong performance on known degradation types, they often exhibit limited generalization capability when encountering unseen degradation patterns or novel corruption intensities. Recently, some works~\cite{Zamir2021RestormerET, Wang_2022_CVPR, NAFNet} show certain generalizability to various degradation types with a unified framework. However, they need to train different models for different degradations, which are not all-in-one solutions as expected. We refer to these methods as multiple-degradation image restoration (MDIR). Subsequently, there have been recent attempts to unify multiple IR tasks within a single framework without the need for retraining. The seminal work~\cite{li2020all} proposed the AiO adverse weather restoration method which utilizes a multi-encoder and decoder architecture and the neural architecture search across task-specific optimized encoders. Moreover, Li \emph{et al.}~\cite{li2022all} first proposed the all-in-one blind image restoration using contrastive learning to extract degradation representations, which can recover images from multiple corruptions without any degradation information in advance. The AiOIR approach aims to recover clean images under various conditions explicitly tailored to address isolated or mixed degradation based on multi-head and multi-tail structures, priors, or pretrained models within a unified framework. It is worth noting that some robust IR models~\cite{wang2021real,wang2024exploiting, yue2023resshift,yu2024scaling, uplavikar2019all, zhang2025uniuir} are still broadly considered part of the AiOIR setting. The AiOIR models present practical advantages such as reduced storage requirements and simplified deployment. However, the principal challenge lies in developing robust architectures capable of effectively addressing diverse degradations and achieving high-quality restorations under varying conditions in real-world scenarios.

\begin{figure*}[t]
	\centering 
	\includegraphics[width=1\linewidth]{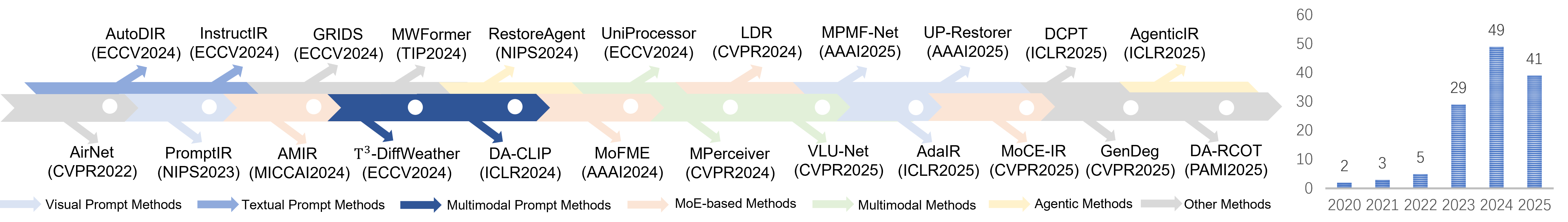}
	\caption{Milestones of AiOIR methods. We simply list their names and venues. The bar chart shows the approximate annual number of papers on AiOIR from 2020 to June 2025.}
	\label{fig:timeline}
\end{figure*}

\subsection{Typical Scenarios}
We categorize AiOIR into several scenarios according to the types of images processed, including natural images, images under adverse weather conditions, underwater images, medical images, document images, hyperspectral images, as well as AI-generated images.

\textbf{Natural Images.} The degradation types commonly encountered in natural images include Gaussian noise, real-world noise, defocus blur, motion blur, low-light conditions, JPEG compression artifacts, mosaic effects, and issues arising from under-display cameras. Various AiOIR approaches have been developed to address these degradations, including techniques described in studies such as~\cite{zhang2023ingredient,potlapalli2024promptir,ma2023prores,cheng2023drm,jiang2023autodir,Yao2023NeuralDR}. These approaches primarily focus on resolving typical problems found in both natural and man-made scenes, enhancing the quality and usability of images affected by these common degradations.

\textbf{Adverse Weather Conditions.}
In more extreme cases, the images requiring restoration may be severely impacted by various adverse weather conditions, such as snowflakes, raindrops, and dense haze. These conditions lead to ill-posed inverse problems that are crucial for applications in autonomous navigation, video retrieval, and outdoor surveillance. The AiOIR field has seen the emergence of many solutions aiming at addressing these challenges, including works like~\cite{Unified,valanarasu2022transweather,ozdenizci2023restoring, zhu2023learning,ye2023adverse,Wang2023GridFormerRD}. Notably, Liu \emph{et al.}~\cite{liu2025clear} proposed multi-weather nighttime image restoration. These solutions are designed to effectively restore visibility and clarity in images affected by harsh environmental conditions.

\textbf{Underwater Images.}
Underwater images suffer from wavelength-dependent light attenuation and scattering caused by marine microorganisms, leading to haze, color casts, low contrast, and loss of fine detail. UniUIR \cite{zhang2025uniuir}, a pioneering framework to tackle such compound distortions in an integrated manner, introduces the Mamba MoE module leveraging DepthAnythingV2~\cite{depthanything_v2} to inject scene depth priors.

\textbf{Medical Images.}
The domain of medical imaging within AiOIR encompasses various types, including clinical computed tomography (CT), magnetic resonance imaging (MRI), and positron emission tomography (PET). Notable approaches in this area include AMIR~\cite{yang2024all} and ProCT~\cite{Ma2023PromptedCT}. AMIR utilizes a task-adaptive routing strategy that achieves state-of-the-art performance across three key medical imaging restoration tasks (\ieno, MRI super-resolution, CT denoising, and PET synthesis). Meanwhile, ProCT introduces an innovative view-aware prompting technique combined with artifact-aware contextual learning. This approach enables universal incomplete-view CT reconstruction and demonstrates seamless adaptability to out-of-domain hybrid CT scenarios.

\textbf{Document Images.}
DocRes~\cite{Zhang2024DocResAG} pioneers the exploration of generalist models for AiOIR of document images. This work addresses five tasks (\ieno, dewarping, deshadowing, appearance enhancement, deblurring, and binarization). DocRes employs a straightforward yet highly effective visual prompting method known as DTSPrompt, which effectively distinguishes between different tasks and accommodates various resolutions. TextDoctor~\cite{lu5108776textdoctor}, utilizing the structure pyramid prediction and patch pyramid diffusion model, exhibits superior performance in capturing the structure of inpainted text.

\textbf{Hyperspectral Images.}
While effective for the RGB image restoration, all-in-one methods struggle with hyperspectral images (HSIs), as the complexity and variability of degradation hinder the generalization of common strategies. PromptHSI~\cite{lee2024prompthsi} integrates frequency-domain priors with the vision-language model driven prompt learning, and introduces a composite degradation dataset for all-in-one HSI restoration. MP-HSIR~\cite{wu2025mp} further introduces a novel multi-prompt framework that integrates spectral, textual, and visual prompts. Unlike PromptHSI, MP-HSIR exploits complementary prompt modalities to better capture diverse degradation patterns and enhance spectral restoration.

\textbf{AI-generated Images.}
With the rise of text-to-image (T2I) models, AI-generated images (AIGIs) often show inconsistent quality, creating an urgent need for models that better match human subjective ratings~\cite{li2024aigiqa}. Li \emph{et al.} proposed Q-Refine~\cite{li2024q}, a quality-aware refiner that enhances both the fidelity and the perceptual quality of AIGIs. G-Refine~\cite{li2024g} further extends Q-Refine by simultaneously refining perceptual details and semantic alignment, achieving better generalization across various T2I models.

\subsection{Task Interactions} 
While the aforementioned taxonomy outlines various scenarios for AiOIR, problems in the real world are inherently complex. It remains crucial to understand the interactions between different restoration tasks, particularly in mixed-degradation scenarios, where multiple types of corruption co-occur (\egno, low-light and blur). These task interactions can be complementary, independent, or even conflicting, significantly affecting the overall restoration quality and the design of unified models. For example:

\begin{itemize}
    \item \emph{Noise and blur}: High ISO or low-light settings often cause both sensor noise and motion blur. Noise may obscure blur kernels, while inaccurate deblurring can amplify noise residuals~\cite{feijoo2024darkir}.
    \item \emph{Rain, haze, and resolution}: In outdoor surveillance, rain streaks and haze not only degrade contrast but also interfere with spatial detail, complicating subsequent super-resolution.
    \item \emph{Color and illumination distortions}: In underwater or adverse weather conditions, color shift, low contrast, and backscatter occur simultaneously, requiring integrated correction of color, visibility, and detail~\cite{schettini2010underwater}.
\end{itemize}

These intertwined degradations present both challenges (task entanglement, ordering sensitivity, domain ambiguity) and opportunities (shared features, multi-task learning, unified representations) for AiOIR models. Understanding these interdependencies is essential for designing robust and generalizable AiOIR systems.

\section{Methodology}
\label{sec:IRmethods}
All-in-one image restoration methods have gained significant attention for their ability to address multiple types of degradation within a unified model. This section analyzes AiOIR from four perspectives: (1) novel and traditional network architectures; (2) diverse learning strategies; (3) advanced techniques such as Prompt Learning, Mixture-of-Experts (MoE), Multimodal Models, and Agentic IR Systems and (4) applications of AiOIR. In Fig.~\ref{fig:timeline}, we present the representative works on AiOIR according to the timeline and method types.

\subsection{Network Design for AiOIR}
\label{sec:network}
\begin{figure}[t]
	\centering \includegraphics[width=\linewidth]{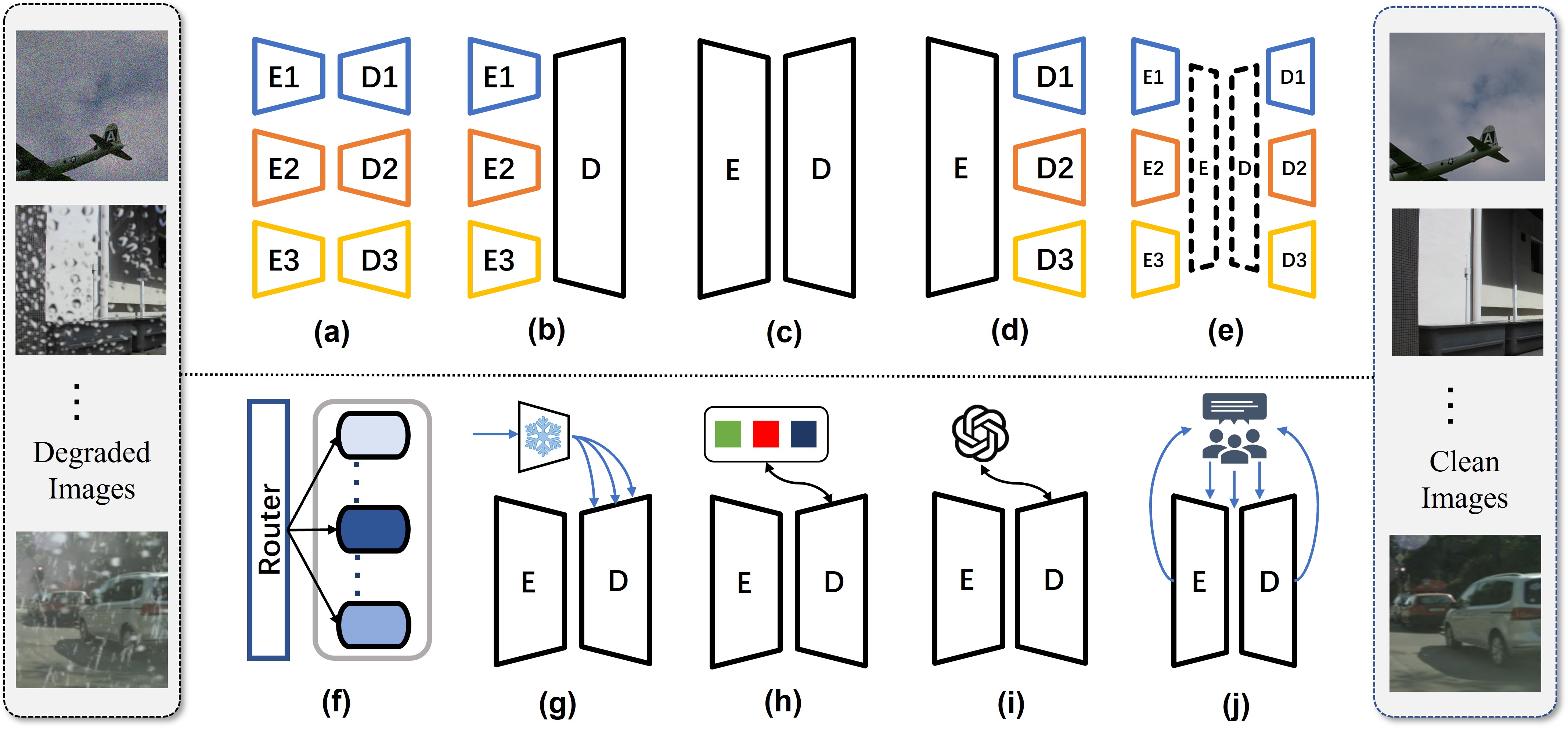}
	\caption{Ten categories of the image restoration method prototype (the last nine are AiOIR methods) (a) independent encoders and decoders. (b) multiple heads and a shared decoder. (c) a single branch end-to-end model. (d) a shared encoder and multiple decoders. (e) a reusable pretrained middle-level backbone. (f) the mixture-of-experts architecture. (g) pretrained large vision-language models to guide image restoration. (h) visual prompts to guide image restoration. (i) textual prompts or instructions to guide image restoration. (j) a question answering paradigm.}
	\label{fig:types}
\end{figure}

Existing AiOIR methods, as a form of multi-task learning (MTL), employ various architectural designs to handle inputs and outputs across multiple tasks, enabling efficient information sharing among them. Despite significant differences in their architectures, AiOIR methods can be broadly categorized into ten representative frameworks, as illustrated in Fig.~\ref{fig:types} and described below.

\textit{(a) Task-Specific Encoders and Decoders:} This straightforward approach assigns a specifically designed encoder-decoder pair to each type of degradation. For example, one encoder-decoder might handle low-light enhancement, while another addresses image denoising. This setup requires prior knowledge of the degradation type to select the appropriate components. However, in real-world scenarios, images often suffer from multiple or unknown degradations, making this approach less practical.

To overcome this limitation, models (b), (c), (d), and beyond aim to handle multiple degradation types within a unified framework without relying on prior knowledge, offering greater flexibility and efficiency.

\textit{(b) Shared Decoder with Multiple Heads:} Models in this category share a common decoder but have multiple heads trained individually for different degradation types. For instance, Li \emph{et al.}~\cite{li2020all} proposed a method for processing various types of adverse weather images using shared weights, but requiring separate training for each degradation type.

\textit{(c) Unified Encoder-Decoder Architecture:} These models use a single encoder-decoder architecture without separate heads or tails, often serving as foundational designs that can be adapted for multi-degradation removal. There are methods like TransWeather~\cite{valanarasu2022transweather}, AirNet~\cite{li2022all}, TANet~\cite{Wang2024TANetTA}, and AdaIR~\cite{cui2024adair}.

\textit{(d) Shared Backbone with Multiple Decoders:} Universal models adopt mixed inputs without any task-specific indicators, using a shared backbone for feature extraction and multiple task-specific decoders. For example, BIDeN~\cite{han2022blind} follows this paradigm. However, this reintroduces the complexity of multiple decoders and requires extensive supervision with degradation labels.

\textit{(e) Pretrained Mid-Level Backbone:} Some models introduce a reusable pretrained transformer backbone with task-specific heads and tails. IPT~\cite{chen2021pre} leverages pretraining to address general noise removal, significantly simplifying the pipeline by utilizing prior knowledge.

\textit{(f) Mixture-of-Experts Architecture:} In MoE-based models, inputs are routed to different experts via a gating mechanism. For instance, MEASNet~\cite{Yu2024MultiExpertAS} considers both pixel-level and global features (including low-frequency and high-frequency components) to select the appropriate expert for IR.

\textit{(g) Pretrained Model Prior:} Models like DINO-IR~\cite{lin2023multi}, Perceive-IR~\cite{Zhang2024PerceiveIRLT}, and DA-CLIP~\cite{Luo2023ControllingVM} utilize frozen vision-language models (\egno, CLIP~\cite{radford2021learning}, DINO~\cite{DINOv1}, and DINO-v2~\cite{Oquab2023DINOv2LR}). These models predict high-quality feature embeddings to enhance the restoration process by leveraging semantic alignment between vision and language.

\textit{(h) Visual Prompting:} These models employ a single encoder-decoder architecture and inject learnable visual prompts at multiple decoding stages to implicitly predict degradation conditions. Examples include works like~\cite{potlapalli2024promptir,li2023prompt,fan2024constyle,zhang2024diff,liu2025dpmambair}. The prompts guide the decoder to adaptively recovering various degraded images, serving as lightweight plug-and-play modules with minimal additional parameters.

\textit{(i) Textual and Multimodal Prompting:} Extending the concept of prompting, models like~\cite{yan2023textual,conde2024high,tian2024instruct} incorporate textual or multimodal prompts. These models allow for natural language instructions or combine visual and textual cues to steer the restoration process, enhancing adaptability to unknown degradations.

\textit{(j) Question-Answering Paradigm:} Finally, models such as PromptGIP~\cite{liu2023unifying} and AutoDIR~\cite{jiang2023autodir} employ a question-answering framework. They empower users to customize image restoration according to their preferences by interpreting user inputs and adjusting the restoration process accordingly.

AiOIR models have evolved from task-specific architectures requiring prior knowledge of degradation types to more flexible and unified frameworks capable of handling multiple degradations without explicit information. Moreover, recent methods increasingly adopt hybrid designs, which we further categorize by their component labels. For example, (g+h) methods such as~\cite{chen2024teaching} combine DepthAnything~\cite{depthanything} with learnable visual prompts; (f+h) methods like~\cite{Yu2024MultiExpertAS} use task‐specific prompts to guide multi-expert selection; (f+i) methods like~\cite{yang2024language} leverage the power of pretrained vision-language models to enrich the diversity of weather-specific knowledge based on an MoE architecture; (h+i) methods, with representative works including~\cite{Zhang2024PerceiveIRLT, Duan2024UniProcessorAT}, embed both visual and textual prompts into a unified framework. With this concise labeling, emerging hybrids can be directly slotted into our taxonomy.

% \begin{figure}[h] \centering \includegraphics[width=1\linewidth]{pics/transfer.png} 
% \vspace{-15pt}
% \caption{Transfer learning scenarios in AiOIR.} \label{fig:transfer} 
% \vspace{-15pt}
% \end{figure}

\subsection{Learning Strategies of AiOIR}
\label{sec:strategies}

Besides network design (as discussed in Sec.~\ref{sec:network}), robust learning strategies are crucial for achieving satisfactory results in AiOIR. In this section, we discuss several promising learning strategies in this field. We begin with continual learning, exploring how it can prevent catastrophic forgetting. Next, we focus on contrastive learning and its application in complex degradation scenarios, emphasizing its role in enhancing the discriminative power of image features. We then discuss zero-shot learning in AiOIR, emphasizing its ability to generalize to unseen degradations. Additionally, we highlight the potential of multi-task learning (MTL) to optimize performance across various degradation tasks, pointing out the importance of addressing task relationships and conflicts. Finally, we introduce the concept of machine unlearning, exploring its potential for privacy protection.

\subsubsection{Continual Learning}

AiOIR methods can be categorized into two learning styles from the perspective of transfer learning~\cite{ruder2019transfer, mao2020surveyselfsupervisedpretrainingsequential}: multi-task learning and sequential learning. MTL involves learning different tasks simultaneously, while sequential learning involves learning different tasks in sequence. In learning models, catastrophic forgetting~\cite{FRENCH1999128} often occurs, where after learning new knowledge, the model almost completely forgets previously learned content. In the field of AiOIR, since a single network is expected to restore images with multiple degradations, the model must learn knowledge related to various degradations, making catastrophic forgetting more likely. To enable the model to incrementally accumulate knowledge and avoid catastrophic forgetting, researchers proposed novel learning strategies, such as review learning~\cite{su2024review} and sequential learning~\cite{kong2401towards}. These methods are inspired by continual learning~\cite{chen2022lifelong, de2021continual} and replace mixed training (mixing datasets with different degradations) with sequential training. It is worth noting that the learning order of multiple tasks is crucial for the quality of IR. In MiOIR~\cite{kong2401towards}, the authors study the effect of training order on the results and point out the impact of multi-task orders. In SimpleIR~\cite{su2024review}, the authors delve into the entropy difference distribution for various IR tasks. They propose determining the training dataset order based on abnormally high loss values and the intrinsic difficulty of tasks, measured by the entropy difference between original and degraded images. Recently, UniCoRN~\cite{mandal2025unicorn}
topologically sorts both single and synthetic multi-degradation datasets based on a parent-child relationship graph and, further sort datasets within a level in the descending order of size to minimize the effect of catastrophic forgetting.

\subsubsection{Contrastive Learning}

One of the significant challenges in IR is effectively handling unseen tasks and degradation types. The vast variability in possible degradations can severely impede a model's generalization capabilities, making it less effective when confronted with new, unseen data. To address this issue, researchers have drawn inspiration from contrastive learning techniques that have proven to be successful in both high-level and low-level tasks~\cite{chen2020simple, he2020momentum, gwu}. Contrastive learning approaches typically serve as an additional form of regularization to improve the generalization of single-task restoration models~\cite{DBLP:conf/cvpr/WuQLZQZ0M21,10176303,gwu,DBLP:conf/cvpr/ZhengZHD023,DBLP:conf/iclr/RanMH0024}. By incorporating contrastive regularization, these methods aim to enhance model performance across a wide array of image restoration applications. In contrastive learning, the definitions of positive and negative samples can be flexibly adjusted, allowing researchers to tailor the learning process to better suit specific tasks and datasets. This flexibility ultimately enhances the model's adaptability and performance in diverse image restoration scenarios~\cite{Zhang2024PerceiveIRLT}. Moreover, contrastive learning-based loss functions have been proposed to obtain discriminative degradation representations in the latent space~\cite{li2022all, zhang2023ingredient, li2023prompt, gwu2025}, further improving the AiOIR model's ability to distinguish between different types of degradations and to generalize to unseen ones.

\subsubsection{Zero-shot Learning}
By restoring images with distortions that are absent in the training data, zero-shot image restoration tackles the challenge of generalizing effectively to unseen degradation types. For instance, methods (\egno, ~\cite{Chung2022ImprovingDM, Chung2022DiffusionPS, Fei2023GenerativeDP, Kawar2022DenoisingDR}) leverage pretrained diffusion models as generative priors, seamlessly integrating degraded images as conditions into the sampling process. In the case of AiOIR, MPerceiver~\cite{ai2024multimodal} demonstrates strong zero-shot and few-shot capabilities across six unseen tasks. Additionally, TAO~\cite{goutest} employs test-time adaptation for AiOIR tasks, achieving results comparable to or better than traditional supervised methods. The core difficulty in these scenarios lies in the distribution shift between training and test data. This requires models to generalize beyond their training experience and adapt to previously unseen degradation types. Test-time adaptation (TTA) (\egno, ~\cite{Shin2022MMTTAMT,Wang2022ContinualTD,Niu2023TowardsST}) has emerged as a key technique in addressing this issue, allowing models to dynamically adjust their parameters during the testing phase to better align with the characteristics of the degraded input images. By enabling the model to restore images with unknown degradations, zero-shot AiOIR aims to provide a more robust and versatile solution, making it a crucial step towards real-world applicability.

\subsubsection{Multi-task Learning}

MTL is a learning paradigm that leverages shared representations and knowledge across tasks, allowing models to learn more efficiently and effectively. By jointly learning from multiple objectives, MTL can improve generalization, reduce overfitting, and achieve better performance on individual tasks. Traditional MTL typically involves a combination of semantic tasks (e.g., classification, detection, segmentation). The tasks exhibit significant semantic differences and also possess distinct output spaces. Typical MTL architectures share a backbone and then branch into task-specific heads, while optimization techniques such as uncertainty weighting, GradNorm, or gradient surgery address conflicts between task gradients~\cite{Yu2020GradientSF, Liu2021ConflictAverseGD, Fifty2021EfficientlyIT}. AiOIR, by contrast, treats each image degradation as a separate task but operates on the same input–output domain (\ieno, pixel-level image reconstruction). Losses remain largely reconstruction-focused (perceptual loss, pixel loss), and task conflicts arise at the gradient level, often manifesting as mutual interference among degradation models. The design of the degradation-aware module is often insufficient to resolve conflicts among different degradations. The optimization process often receives less attention, leading to oversight of the complex relationships and potential conflicts between multiple degradations in mixed training scenarios. Unlike unified models (\egno, ~\cite{potlapalli2024promptir, kong2401towards, valanarasu2022transweather, li2022all}) that employ mixed training, some novel studies address AiOIR from the perspective of MTL to resolve inconsistencies and conflicts among different degradations, by treating each degradation as an independent task. By focusing on the optimization process and the interactions between tasks, these methods aim to mitigate conflicts and enhance overall performance. We can broadly classify AiOIR methods in MTL into two types: task grouping and task balancing.

\textbf{Task Grouping.} A notable example is GRIDS~\cite{Cao2024GRIDSGM}, which enhances MTL by strategically dividing tasks into optimal groups based on their correlations. Tasks that are highly related are grouped together, allowing for more effective training. GRIDS introduces an adaptive model selection mechanism that automatically identifies the most suitable task group during testing. This approach leverages the benefits of group training, ultimately improving overall performance by ensuring that related tasks are processed in a complementary manner.

\textbf{Task Balancing.} Conversely, Wu \emph{et al.}~\cite{wu2024harmony} proposed a straightforward yet effective loss function that incorporates task-specific reweighting. This method dynamically balances the contributions of individual tasks, fostering harmony among diverse tasks. By adjusting the weight of each task based on specific characteristics and performance, this approach mitigates conflicts and enhances the overall effectiveness. Subsequently, Wu \emph{et al.} proposed TUR~\cite{wu2025debiased}, a task-aware optimization strategy that introduces adaptive task-specific regularization for AiOIR.

\subsubsection{Machine Unlearning}

Privacy protection is a critical concern in AI, especially as models are deployed in sensitive applications. Machine unlearning~\cite{Cao2015TowardsMS} offers a way to remove the influence of private data from trained models, enabling them to behave as if such data were never used. Existing methods fall into two categories: exact unlearning~\cite{Bourtoule2019MachineU}, which seeks to completely eliminate the impact of specific data points, and approximate unlearning~\cite{Nguyen2020VariationalBU}, which aims to diminish the influence of such data to a certain degree. Although unlearning has been studied in classification and federated learning, its use in AiOIR remains underexplored.

To fill this gap, Su \emph{et al.}~\cite{Su2024AccurateFF} defined a scenario in which certain types of degradation are treated as private information that must be effectively forgotten by the model. Importantly, this process is designed to preserve the model's performance on other types of degradations. They introduced instance-wise unlearning, which leveraged adversarial examples in conjunction with gradient ascent methods. This approach not only enhances the AiOIR model's ability to forget specific data but also maintains its robustness across various image restoration tasks.

\subsection{Key Improvements of AiOIR}
\label{sec:keyimprove}
\begin{figure*}[htp]
	\centering 
	\includegraphics[width=1\linewidth]{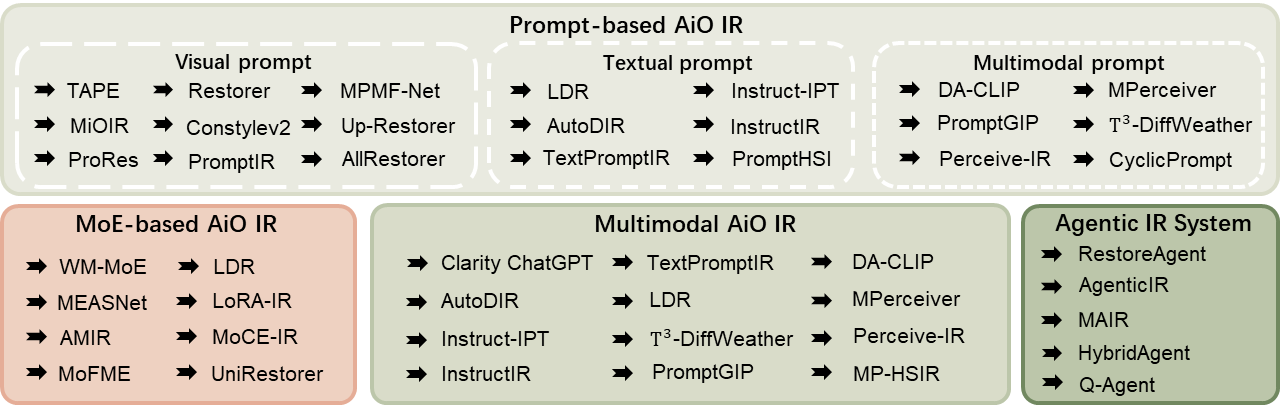}
	\caption{The overview of AiOIR methods. This figure categorizes AiOIR into four types based on their key improvements, namely the Prompt-based models, MoE-based models, Multimodal models and Agentic IR systems.}
	\label{fig:methods}
\end{figure*}

In addition to the network design (Sec.~\ref{sec:network}) and learning strategies (Sec.~\ref{sec:strategies}), other key techniques further enhance AiOIR models. We categorize these improvements into four areas: Prompt Learning, Mixture-of-Experts (MoE), Multimodal Model and Agentic IR System, which correspond to Sec.~\ref{sec:pl}, Sec.~\ref{sec:moe}, Sec.~\ref{sec:mm}, Sec.~\ref{sec:agent}, respectively. We also highlight additional techniques such as deep unfolding methods, mask image modeling, hypernetworks, \emph{etc}. At last, we discuss AiOIR for downstream tasks and outline real-world applications. For clarity, representative works for each category are summarized in Fig.~\ref{fig:methods}.

\subsubsection{Prompt Learning}
\label{sec:pl}
Prompt learning, initially successful in natural language processing (NLP)~\cite{brown2020language,gao2020making,liu2021p}, aims to tap into the knowledge that is inherent in the language model by providing instructions or relevant information. Inspired by the ability to effectively model task-specific context, prompts have been used for fine-tuning for vision tasks~\cite{jia2022visual,khattak2023maple}. To be specific, unlike single-task image restoration, learnable prompts enable better parameter-efficient adaptation of models facing multiple degradations. Recently, various prompts have been explored in AiOIR, as an adaptive lightweight module to encode degradation representation in the network. The core idea is to enable pretrained models to better understand and perform downstream tasks by constructing visual, textual or multimodal prompts, which are as follows.

\textbf{Visual Prompting.}
Visual prompting has gained widespread attention for tackling both high-level and low-level vision tasks~\cite{bar2022visual, chen2022unified, wen2025multi}. In AiOIR, prompt learning enables models to adaptively select task-specific prompts, achieving strong performance across diverse degradations. For example, AirNet~\cite{li2022all} leverages degradation-aware features, while Transweather~\cite{valanarasu2022transweather} uses query-based guidance. This growing trend highlights the increasing integration of prompt learning in AiOIR research.

\begin{figure}[!htbp]
    \centering
    \includegraphics[width=1\linewidth]{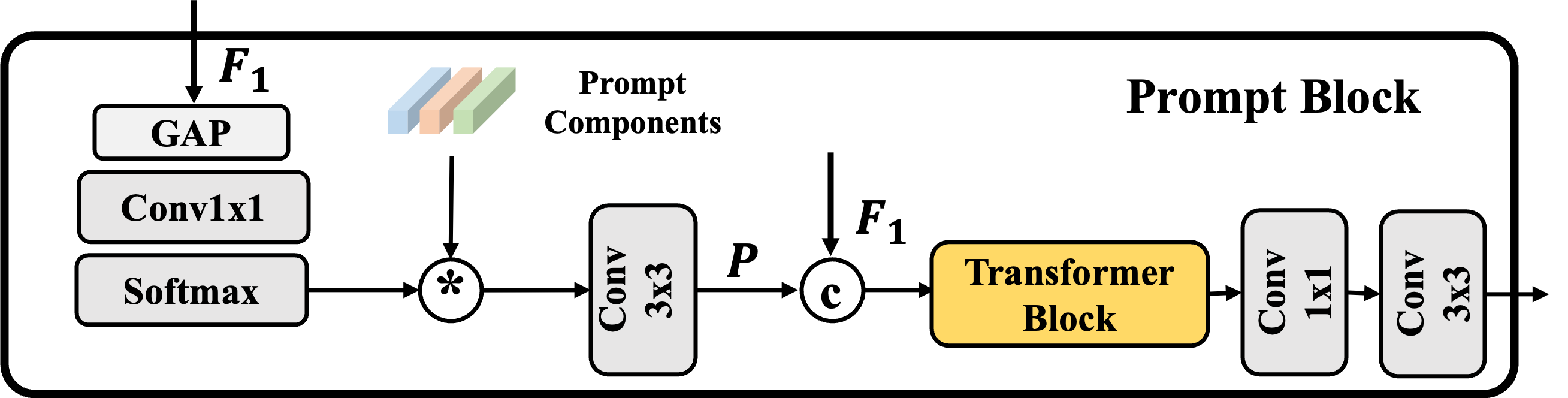}
    \caption{Illustration of the prompt block in AiOIR.}
    \label{fig:prompt}
\end{figure}

PromptIR~\cite{potlapalli2024promptir} is one of the most representative works, integrating a prompt block into the U-Net architecture to enhance Restormer~\cite{Zamir2021RestormerET}. The prompts serve as an adaptive, lightweight plug-and-play module as shown in Fig.~\ref{fig:prompt}, enabling multi-scale degradation-aware guidance across layers. Extending this idea, ProRes~\cite{ma2023prores} embeds visual prompts directly into the input image. PromptGIP~\cite{liu2023unifying} proposes a training method similar to masked autoencoding~\cite{He2021MaskedAA}, where certain portions of question and answer images are randomly masked. This prompts the model to reconstruct these patches from the unmasked areas. PIP~\cite{li2023prompt} further introduces a novel Prompt-In-Prompt learning framework for AiOIR, which employs two innovative prompts: a degradation-aware prompt and a basic restoration prompt. To improve task-specific representation, Wen \emph{et al.}~\cite{wen2025multi} propose multi-axis prompt and multi-dimension fusion network for all-in-one traffic weather removal by incorporating prompts from three axes. CPL~\cite{gwu2025} establishes a new paradigm that extends contrastive learning principles from feature-level discrimination to the previously unexplored domain of prompt-level alignment. Recently, Wu \emph{et al.}~\cite{wu2025learning} introduce prompt distribution learning, which models prompts as distributions rather than fixed parameters.

\textbf{Textual Prompting.}
While earlier AiOIR studies focused on learnable visual prompts, their effectiveness was limited by the semantic gap~\cite{Pang2019TowardsBS}, making it difficult to accurately identify degradation types. To address this, recent works~\cite{tian2024instruct, conde2024high, jiang2023autodir, yan2023textual, yang2024language, wei2023clarity} explore textual prompts for more precise and user-friendly guidance. For instance, TextPromptIR~\cite{yan2023textual} leverages a fine-tuned BERT model to interpret task-specific instructions, enabling semantic understanding of degradations. Text prompts allow users to describe degradations in natural language, enhancing model adaptability and accessibility. Building on this, Clarity ChatGPT~\cite{wei2023clarity} and AutoDIR~\cite{jiang2023autodir} combine large language models with visual features to support intuitive and interactive IR, even in unseen tasks. Extending to hyperspectral image restoration, PromptHSI~\cite{lee2024prompthsi} introduces text-guided, frequency-aware modulation, showcasing the growing potential of language-driven AiOIR.

\textbf{Multimodal Prompting.}
Vision-language models (VLMs) have shown strong potential in AiOIR by leveraging aligned visual and textual representations. Contrastive learning allows VLMs to train image and text encoders, thereby enabling prompts that integrate text for holistic semantics and vision for fine-grained details. Building on this, MPerceiver~\cite{ai2024multimodal} utilizes stable diffusion priors and CLIP-based dynamic prompts to adaptively handle diverse degradations. DA-CLIP~\cite{radford2021learning} introduces an image controller to align degraded inputs with high-quality CLIP embeddings. CyclicPrompt~\cite{liao2025prompt} fuses CLIP-derived weather knowledge, conditional vectors, and textual prompts for context-aware restoration. MP-HSIR~\cite{wu2025mp} focuses on HSI restoration by integrating spectral, visual, and textual prompts. Recently, VLU-Net~\cite{zeng2025vision} leverages CLIP-derived high-dimensional features for automatic selection of degradation-aware keys in deep unfolding networks. These works highlight multimodal prompting in enhancing adaptability and generalization in AiOIR.

\subsubsection{Mixture-of-Experts}
\label{sec:moe}
The mixture-of-experts (MoE) framework, first proposed in “Adaptive Mixture of Local Experts”~\cite{6797059}, employs multiple specialized networks alongside a gating network that dynamically assigns weights to each expert based on the input. This architecture allows experts to focus on different data distributions, enhancing model specialization and performance in complex tasks. Researchers in AiOIR have observed that IR model parameters tend to be degradation-specific. For example, parameters related to one type of degradation are typically inactive when dealing with other types of degradation, and zeroing out these unrelated parameters has little impact on image restoration quality, as illustrated in LDR~\cite{yang2024language}. This observation aligns with the concept of conditional computation in MoEs, where sparsity plays a key role. Applying MoEs to the field of AiOIR could lead to several improvements, including faster pretraining compared to dense models and quicker inference with the same number of parameters.

% \begin{figure}[ht]
%     \centering
%     \includegraphics[width=1\linewidth]{pics/moe.png}
%     \caption{Schematic illustration of the Mixture-of-Experts in AiOIR.}
%     \label{fig:moe}
% \end{figure}

Recent studies have extended the MoE paradigm to handle weather-related degradations using both language-driven and weather-aware routing mechanisms. For example, Yang \emph{et al.}~\cite{yang2024language} generated degradation priors based on textual descriptions of weather conditions, which are then used to guide the adaptive selection of restoration experts. Similarly, WM-MoE~\cite{luo2023wm} introduces a weather-aware routing mechanism to direct image tokens to specialized experts and employs multi-scale experts to handle diverse weather conditions. Both WM-MoE and MoFME~\cite{Zhang2023EfficientDM} enhance IR tasks and downstream tasks like segmentation, while significantly reducing model parameters and inference time. In contrast, MEASNet~\cite{Yu2024MultiExpertAS} presents a novel multi-expert adaptive selection mechanism. This mechanism leverages both local and global image features to select the most appropriate expert models, effectively balancing task-specific demands and promoting resource sharing. To further adapt to varying complexities, MoCE-IR~\cite{zamfir2024complexity} introduces complexity experts with adaptive variable-size computational units that use a spring-inspired force mechanism to dynamically allocate experts. Building on similar principles, UniRestorer~\cite{lin2024unirestorer} constructs a multi-granularity degradation set and a multi-granularity MoE restoration model. LoRA-IR~\cite{ai2024taming} leverages a novel mixture of low-rank experts structure, combined with a CLIP-based degradation-guided router. Recently, M2Restore~\cite{wang2025m2restore} introduces a novel CLIP-guided MoE-based
Mamba-CNN framework. Overall, the sparse activation nature of MoE architectures enables more resource-aware AiOIR models with improved inference efficiency and generalizability.

\subsubsection{Multimodal Model}
\label{sec:mm}
Multimodal learning has become increasingly pivotal in the domain of low-level vision~\cite{pu2025lumina}, enabling the enrichment of visual understanding through the integration of diverse information sources. The fundamental objective of multimodal tasks is to learn potential feature representations from a multitude of modalities, such as textual captions and visual images, RGB images with supplementary components like depth or thermal images, and various forms of medical imaging data. Similarly, multimodal models in IR harness data from multiple sources to improve the fidelity and robustness of restored images. These models integrate complementary modalities to address limitations inherent in single-modality approaches, particularly in scenarios involving complex degradation. By leveraging different types of information, multimodal models enhance the structural details, texture, and overall quality of the restored image.

However, multimodal models also introduce challenges, including the increased computational complexity of handling diverse data streams and the requirement for well-aligned multimodal datasets. Additionally, the process of effectively fusing different types of data, which may have varying resolutions and characteristics, remains a significant technical hurdle. Here, we summarize various approaches to multimodal AiOIR methods (\egno, Clarity-ChatGPT~\cite{wei2023clarity}, AutoDIR~\cite{jiang2023autodir}, Instruct-IPT~\cite{tian2024instruct}, InstructIR~\cite{conde2024high}, \emph{etc}). These involve the continuous guidance of IR via human language instructions, as well as the use of multimodal prompts for AiOIR, as previously mentioned. Clarity-ChatGPT is notable as the system that bridges adaptive image processing with interactive user feedback, innovatively integrating large language and visual models. AutoDIR automatically detects and restores images with multiple unknown degradations by the semantic-agnostic blind image quality assessment (SA-BIQA). InstructIR~\cite{conde2024high} trains models using common image datasets and prompts generated by GPT-4, noting that this generalizes to human-written instructions. Moreover, CMAWRNet~\cite{frants2025cmawrnet} leverages structure and texture decomposition images, and $\mathrm{T}^{3}\text{-}\mathrm{DiffWeather}$~\cite{chen2024teaching}  uses DepthAnything~\cite{depthanything} features as a constraint for general prompts. Recently, Perceive-IR~\cite{Zhang2024PerceiveIRLT} leverages multi-level quality-driven prompts categorized into three tiers of quality by image-text pairs to perceive severity levels.

\subsubsection{Agentic IR System}
\label{sec:agent}
\begin{figure}[!htbp]
    \centering
    \includegraphics[width=1\linewidth]{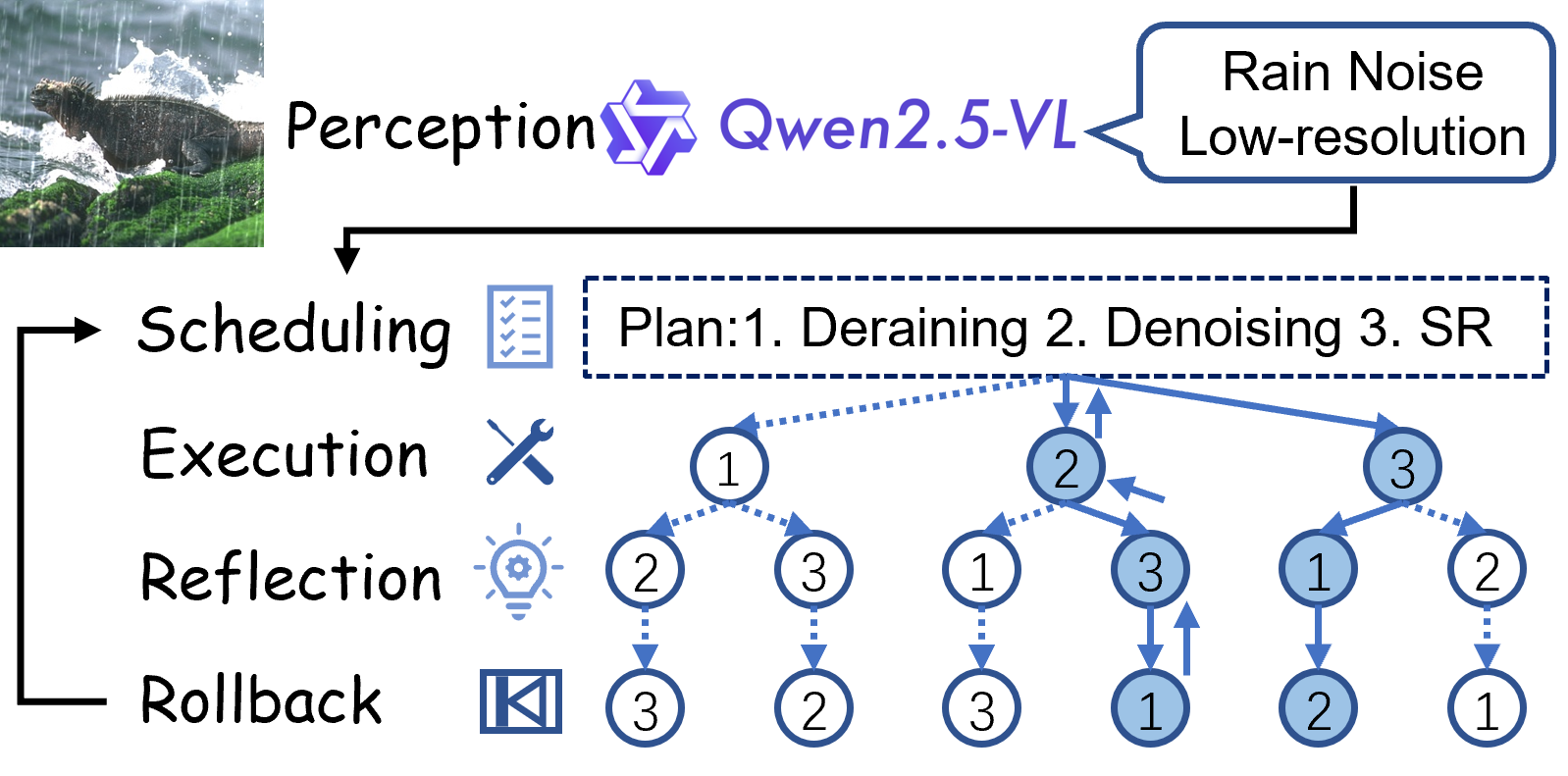}
    \caption{Illustration of Agentic IR System.}
    \label{fig:agen}
\end{figure}
The rise of Agentic IR Systems marks a paradigm shift in handling complex real-world degradations, positioning them as a natural extension of AiOIR~\cite{zhu2024intelligent, zhou2025q}. Powered by multimodal large language models (MLLMs), these systems enable autonomous analysis, dynamic planning, and adaptive model selection as shown in Fig.~\ref{fig:agen}. For instance, RestoreAgent~\cite{chen2024restoreagent} introduces modular restoration guided by degradation-aware planning, while CoR~\cite{cao2024chain} proposes a step-by-step chain-of-restoration strategy, albeit more suited for non-blind scenarios. Notably, AgenticIR~\cite{zhu2024intelligent} mimics human workflows through perception, scheduling, and adaptive execution using LLM-VLM collaboration. MAIR~\cite{jiang2025multi} structures degradation handling via a three-stage framework based on real-world priors, reducing the search space through expert agents. HybridAgent~\cite{li2025hybrid} further enhances efficiency by combining fast and slow agents, enabling ambiguity resolution while minimizing error propagation. More recently, Q-Agent~\cite{zhou2025q} first fine-tunes MLLMs, and uses chain-of-thought to decompose multi-degradation perception. Collectively, these agentic approaches integrate reasoning, modularity, and prior knowledge to boost robustness and flexibility in multi-degradation restoration.

\subsubsection{Other Methods}

In addition to highlighting four key improvements, we also review other methods for AiOIR. Some models benefit from iterative algorithms in deep unfolding networks (DUNs), and integrating network design with pretrained mask image modeling (MIM) also holds significant potential. Meanwhile, hypernetworks with dynamically generated parameters are also highly suitable for AiOIR. The exploration of visual autoregressive modeling (VAR) as a novel paradigm for low-level vision tasks has recently extended to the field of AiOIR. The details are as follows.

\textbf{Deep Unfolding Methods.}
Under the framework of optimization theory and deep learning, DUNs bridge interpretability and data-driven flexibility by translating iterative solvers into end-to-end trainable architectures. Zhang \emph{et al.}~\cite{Zhang2017LearningDC} proposed a DUN framework that integrates CNNs with model-based priors for task-specific IR. Given a degradation model, the clean image is estimated by minimizing an energy function. The half-quadratic splitting (HQS) algorithm~\cite{Geman1995NonlinearIR} decomposes the optimization into two subproblems: one for the data fidelity term and the other for the prior. These subproblems are alternately solved, with the data term handled via least squares and the prior term approximated using CNNs. Building on this principle, DRM-IR~\cite{cheng2023drm} enhances the flexibility of AiO scenarios by introducing a reference-based and task-adaptive modeling paradigm. It establishes an efficient AiOIR framework that jointly models task-adaptive degradation and model-based restoration. Recently, Up-restore~\cite{liu2025up} employs a prompt-driven unrolling scheme inspired by ADMM and grounded in MAP estimation, effectively combining data-driven learning with unified degradation modeling. Additionally, VLU-Net~\cite{zeng2025vision} leverages an automatic gradient estimation strategy based on VLMs, integrating multi-level information to improve feature preservation. These works pave the way for more flexible, efficient, and robust solutions for tackling complex, real-world AiOIR challenges.

\begin{table*}[htp]
  \centering
  \caption{Summary of used datasets in AiOIR tasks.}
    \resizebox{\textwidth}{!}{\begin{tabular}{clccccc}
    \Xhline{1.1pt}
    \multicolumn{1}{c}{\textbf{Task}}  & \multicolumn{1}{l}{\textbf{Dataset}} & \textbf{Year}  & \textbf{Training} & \textbf{Testing} & \textbf{Short Description} & \textbf{Type} \\ \Xhline{1.1pt}
    \multirow{3}[0]{*}{Image Deblurring} 
          & GoPro~\cite{Gopro} & 2017  & 2103  & 1111   & Blurred images at 1280x720 & real\\
          & HIDE~\cite{HIDE}  & 2019  & 6397  & 2025   & Blurry and sharp image pairs & real\\
          & RealBlur~\cite{Real-blur} & 2020  & 3758  & 980    & 182 different scenes & real \\
    \hline
    \multirow{6}[0]{*}{Image Denoising}
          & Kodak~\cite{franzen1999kodak}& 1999  & -  &24  & Lossless true color image suite & real\\
          & Urban100~\cite{Huang-CVPR-2015}& 2015  & -  &100  & 100 images of urban scenes & real\\
          & WED~\cite{ma2016waterloo}& 2016  & 4744  &-  &  Images collected from Internet & real\\

          & BSD400~\cite{arbelaez2010contour} & 2010  & 400  &-  & 400 clean natural images & real\\
          & CBSD68~\cite{BSD100} & 2001  & - & 68   & Images with different noisy levels & real\\
          & McMaster~\cite{zhang2011color-mcmaster} & 2011  & - & 18   & Crop size=500x500 & real\\
    \hline
    \multirow{5}[0]{*}{Image Super-resolution} 
          & DIV2K~\cite{DIV2K} & 2017  & 800   & 100  & 2K resolutions & real \\
          & Flickr2K~\cite{wang2019flickr1024} & 2017  & 2650  & -     & 2K resolutions & real\\
          % & Set5~\cite{set5} & 2012  & -     & 5    & Classic 5 images & real\\
          & Set14~\cite{Set14} & 2012  & -     & 14    & Classic 14 images & real\\
          & BSD100~\cite{BSD100} & 2001  & -     & 100   & Objects, Natural images & real\\
          % & Manga109~\cite{manga109} & 2015  & -     & 109  & 109 manga volumes & synthetic\\
          & Urban100~\cite{Huang-CVPR-2015} & 2015  & -     & 100   & 100 urban scenes & real\\
    \hline
    \multirow{2}[0]{*}{Shadow Removal} 
          & ISTD~\cite{ISTD}  & 2018  & 1330  & 540    & 135 scenes with shadow mask images & synthetic\\
          & SRD~\cite{SRD}   & 2017  & 2680  & 408    & Large scale dataset for shadow removal & real\\
    \hline
    \multirow{5}[0]{*}{Image Desnowing} 
          & Snow100k~\cite{snow100k} & 2017  & 50000 & 50000  &  Including 1369 realistic snowy images & real and synthetic\\
          & CSD~\cite{chen2021all} & 2021 & 2000&-&A large-scale dataset called Comprehensive Snow Dataset&synthetic\\
          & REVIDE~\cite{zhang2021learning} & 2021 & 1698&284&Real-world Video Dehazing& real\\          
          & RealSnow+~\cite{zhu2023learning} & 2023 & 1650 & 240 & Containing various resolutions&real\\
          % & KITTI-snow~\cite{wang2023weatherdepth} & 2023 & 50 & - & Two intensity levels: severe and extremely severe&synthetic\\
          & WeatherStream~\cite{Zhang2023WeatherStreamLT} & 2023 & 176100 & 11400 &  Three weather conditions, \ieno, rain, snow, and fog &real\\
    \hline
    \multirow{5}[0]{*}{Image Deraining} 
          & RainDrop~\cite{rain-drop} & 2018  & 1119  & -     & Various scenes and raindrops & real\\
          & Outdoor-Rain~\cite{outdoor-rain} & 2019  & 9000  & 1500   & Degraded by both fog and rain streak & synthetic\\
          & SPA~\cite{wang2019SPA-data} & 2019  & 295000 & 1000   & Various natural rain scenes &real \\
          % & Rain-200~\cite{yang2017deep-rain100} & 2017  & 1800    & 100    & Five rain streaks& synthetic \\
          & Rain1400~\cite{fu2017removing} & 2017 &12600 & 1400& Diverse rainfall directions and levels& synthetic\\
          & Rain100L/H~\cite{yang2017deep-rain100} & 2017  & 200    & 100    & Five rain streaks & synthetic\\
    \hline
    \multirow{4}[0]{*}{Image Dehazing} 
          & Dense-Haze~\cite{ancuti2019dense-densehaze} & 2019  & 33&-  & Out-door hazy scenes & real\\
          % & RESIDE~\cite{li2019benchmarking} & 2019  & 423950 & 5342   & Real-world hazy images & synthetic and real\\
          & RESIDE-OTS~\cite{li2019benchmarking} & 2019 & 313950& - & Outdoor training set & real\\
          & RESIDE-ITS~\cite{li2019benchmarking} & 2019 & 110000& - & Indoor training set & synthetic\\
          & RESIDE-HSTS~\cite{li2019benchmarking} & 2019 & -& 20 & Hybrid subjective testing set & synthetic and real\\
          % & RESIDE-SOTS~\cite{li2019benchmarking} & 2019 & -& 1000 & Synthetic objective testing set & synthetic\\
    \hline
    \multirow{1}[0]{*}{Low-light Image Enhancement} 
          & LOL~\cite{wei2018deep} & 2018 & 485 & 15 & Low- and normal-light images at 400x600 & real\\
          % & LLIV-Phone~\cite{LoLi} & 2021 & 45148 & - & Taken by 18 different phones' cameras & real\\
          % & NPE~\cite{NPE} & 2013 & 84 & - & Unpaired images & real\\
          % & MEF~\cite{MEF} & 2015 & 17 & - & From the Internet and from existing datasets  & real\\
    \Xhline{1.1pt}
    \end{tabular}}
  \label{tab:datasets}
\end{table*}

\textbf{Large-scale Vision Models.}
Recent works have demonstrated the potential of pretrained vision-language models (VLMs) to enhance downstream tasks using generic visual and text representations~\cite{radford2021learning,Jia2021ScalingUV,Li2022BLIPBL}. A classic VLM typically consists of a text encoder and an image encoder, learning aligned multimodal features from image-text pairs through contrastive learning~\cite{radford2021learning}. BLIP~\cite{Li2022BLIPBL} improves this by eliminating noisy web data with synthetic captions. CLIP~\cite{radford2021learning} has demonstrated effective semantic alignment between vision and language, aiding numerous downstream tasks. In parallel, self-supervised models such as DINO~\cite{DINOv1} and DINO-v2~\cite{Oquab2023DINOv2LR} have shown strong performance without requiring labeled data, further expanding the applicability of pretrained representations. VLMs have also gained significant traction in the domain of AiOIR~\cite{shao2024adaptive, Zhang2024PerceiveIRLT, Luo2023ControllingVM, lin2023multi}. Perceive-IR~\cite{Zhang2024PerceiveIRLT} and DINO-IR~\cite{lin2023multi} utilize the semantic prior knowledge and structural information mined by a DINO-based guidance module to enhance the restoration process. DA-CLIP~\cite{Luo2023ControllingVM} trains an additional controller that adapts the fixed CLIP image encoder to predict high-quality feature embeddings. AWRaCLe~\cite{rajagopalan2025awracle} employs visual in-context learning, incorporating the feature differences between the clean and degraded image pairs extracted by CLIP at the decoder.

\textbf{Mask Image Modeling.}
Mask image modeling (MIM) (\egno, ~\cite{Xie2021SimMIMAS,He2021MaskedAA, Wang2022ImagesSI}) is a technique that involves training models to predict masked portions of images based on their surrounding context inspired by mask language modeling~\cite{Radford2018ImprovingLU,Devlin2019BERTPO}. This self-supervised approach reconstructs missing image regions, enhancing its visual representation learning. MIM has demonstrated effectiveness in visual tasks such as classification and segmentation. The MAE~\cite{He2021MaskedAA} framework effectively employs MIM to predict hidden tokens, showcasing impressive performance and generalization capabilities across a range of downstream tasks. Qin \emph{et al.}~\cite{Qin2024RestoreAW} are the pioneers in introducing MIM into AiOIR, shifting the focus from degradation‐type discrimination to intrinsic image information recovery. They proposed mask attribute conductance (MAC) to analyze each layer’s contribution to resolving the input integrity gap and rank layers in descending order. DyNet ~\cite{Dudhane2024DynamicPT} trains two variants (small and large) concurrently to reconstruct the clean image from masked degraded inputs. Notably, both variants share weights via an intra-network block-repetition scheme, reducing GPU hours by 50\%. Most recently, Cat-AIR ~\cite{jiang2025cat} introduces a content-aware and task-aware router that dynamically applies self-attention to complex regions and convolution to simpler ones, using learned masks to distinguish hard from easy patches within the feature map.

\textbf{Hypernetworks.}
Hypernetworks~\cite{ha2016hypernetworks} are a class of neural networks designed to generate parameters for other networks. This enables the network to dynamically adapt to varying input patterns, enhancing its flexibility. HAIR~\cite{Cao2024HAIRHA} introduces data-conditioned hypernetworks into the AiOIR by incorporating the plug-and-play modules into Restormer~\cite{Zamir2021RestormerET} to dynamically generate transformer block weights. Zhu \emph{et al.}~\cite{zhu2024mwformer} propose MWFormer for multi-weather IR. A hypernetwork is employed to extract content-independent, weather-aware features. It also generates the weights and biases for feature vectors, along with parameters for depthwise convolutions and attention modules. Furthermore, it produces weather-informed feature vectors for both weather-type identification and the guidance of pretrained weather-specific models.

\textbf{Visual Autoregressive.}
Tian~\emph{et al.}~\cite{tian2024visual} proposed visual autoregressive modeling (VAR), a new generation paradigm that redefines the autoregressive learning and achieves efficiency and generative performance superior to stable diffusion \cite{rombach2021highresolution}. For AiOIR, Varformer~\cite{wang2024varformer} explores the multi-scale representations learned by VAR and reveals its endogenous distribution alignment priors, which transition from global color information to fine-grained details. Unlike Varformer, which utilizes intermediate VAR features to guide a separate non‐generative network, RestorerVAR~\cite{rajagopalan2025restorevarvisualautoregressivegeneration} is generative. It fully exploits the strong priors of the pretrained VAR model by introducing degraded‐image conditioning through cross‐attention at each transformer block.

\subsection{Applications of AiOIR}
\label{sec:apply}

Beyond perceptual enhancement, AiOIR methods can also benefit downstream tasks. For example, UniRestore~\cite{chen2025unirestore} trains a task-aware adapter to align encoder and latent features with various downstream objectives (\egno, classification, semantic segmentation). Similarly, MoFME~\cite{Zhang2023EfficientDM} demonstrates robust performance for both upstream and downstream tasks under adverse weather. Beyond these, AiOIR has promising applications in tasks like depth estimation, remote sensing, medical imaging, and autonomous driving, highlighting the need for further research on practical deployment.

\section{Experiments}
\label{sec:exp}
To facilitate a comprehensive comparison of AiOIR methods, we summarize key datasets and metrics for AiOIR tasks, then compare representative benchmarks under five common experimental setting. This section enables a thorough evaluation of AiOIR methods.

\subsection{Datasets and Evaluation Metrics}
\label{sec:datasets}
 
\textbf{Datasets.}
A wide range of datasets are available for AiOIR, varying significantly in terms of image quantity, quality, resolution, and diversity. 
% Some datasets offer paired input and target images, while others provide only ground-truth images. In the latter case, LQ images are generally manually generated. For example, BSD~\cite{arbelaez2010contour} is a classical dataset for image denoising and super-resolution. BSD100 is a classical image dataset containing 100 test images proposed by Martin \emph{et al.}. The dataset is composed of a large variety of images ranging from natural images to object-specific images such as plants, people, food, \emph{etc}. Notably, the real-world distortions are usually blind or unknown, where the distributions are different from the simple synthetic distortions.
Tab. \ref{tab:datasets} summarizes the datasets used for different AiOIR tasks. Recently, FoundIR~\cite{li2024foundir} contributes a million-scale high-quality dataset for IR foundational model, highlighting the potential of AiOIR in real-world scenarios.

\textbf{Evaluation Metrics.}
The performance of AiOIR methods is typically evaluated using three aspects of metrics: Distortion metrics(\egno, PSNR, SSIM~\cite{SSIM}) refer to the relationship between the restored image and the original image. Perceptual metrics(\egno, {FID}~\cite{FID}, LPIPS~\cite{LPIPS}) evaluate how much the image appears like a natural image. No-reference metrics(\egno, {NIQE}~\cite{mittal2012makingNIQE}, BRISQUE~\cite{Mittal2012NoReferenceIQ}) are commonly based on estimating deviations from natural image statistics. In addition, there are several metrics that play a crucial role in comparing the performances of different AiOIR methods, including IL-NIQE~\cite{IL-NIQE}, NIMA~\cite{NIMA}, CLIP-IQA~\cite{CLIP-IQA}, LOE~\cite{LOE}, Consistency~\cite{Con}, PI~\cite{mittal2012makingNIQE}, and MUSIQ~\cite{MUSIQ}.

\begin{table*}[htp]
\centering
\caption{Performance Comparisons of AiOIR Models on Three Challenging Datasets.
}\label{tab:results1}
    \renewcommand\arraystretch{1.15}
\resizebox{\textwidth}{!}{
  \begin{tabular}{l|lllcccccccl}
    \Xhline{1.1pt}
    \multirow{2}{*}{\textbf{}} & \multirow{2}{*}{\textbf{Method}} &\multirow{2}{*}{\textbf{Venue \& Year}}&\multirow{2}{*}{\textbf{Params}} & \textbf{Dehazing} & \textbf{Deraining} &  \multicolumn{3}{c}{\textbf{Denoising} on BSD68 dataset~\cite{BSD100}} &\multirow{2}{*}{\textbf{Average}}&\multirow{2}{*}{\textbf{Approach}}\\ 
     & & & & SOTS~\cite{li2019benchmarking}& Rain100L~\cite{yang2017deep-rain100}& $\sigma = 15$ & $\sigma = 25$ & $\sigma = 50$&&& \\
    \Xhline{1.1pt}
    \multirow{4}{*}{\rotatebox{90}{Single}} 
    & LPN~\cite{LPN} &CVPR'19&3M& 20.84/0.828 & 24.88/0.784 & 26.47/0.778 & 24.77/0.748 & 21.26/0.552 &23.64/0.738&Specific\\ \cline{2-12}
    &ADFNet~\cite{ADFNet} &AAAI'23&8M& 28.13/0.961 & 34.24/0.965 & 33.76/0.929 & 30.83/0.871 & 27.75/0.793 &30.94/0.904&Specific\\ \cline{2-12}
    &DehazeFormer~\cite{DehazeFormer} &TIP'23&25M& 29.58/0.970 & 35.37/0.969 & 33.01/0.914 & 30.14/0.858 & 27.37/0.779 &31.09/0.898&Specific\\ \cline{2-12}
    &DRSformer~\cite{DRSformer} &CVPR'23&34M& 29.02/0.968 & 35.89/0.970 & 33.28/0.921 & 30.55/0.862 & 27.58/0.786 &31.26/0.902&Specific\\ \hline \hline
    \multirow{5}{*}{\rotatebox{90}{Multiple}} 
    &MPRNet~\cite{MPRNet} &CVPR'21&16M& 28.00/0.958 & 33.86/0.958 & 33.27/0.920 & 30.76/0.871 & 27.29/0.761 &30.63/0.894&General\\ \cline{2-12}
    &Restormer~\cite{Zamir2021RestormerET} &CVPR'22&26M& 27.78/0.958 & 33.78/0.958 & 33.72/0.865 & 30.67/0.865 & 27.63/0.792 &30.75/0.901&General\\ \cline{2-12}
    &NAFNet~\cite{NAFNet} &ECCV'22&17M& 24.11/0.960 & 33.64/0.956 & 33.18/0.918 & 30.47/0.865 & 27.12/0.754 &29.67/0.844&General\\ \cline{2-12}
    &FSNet~\cite{FSNet} &TPAMI'23&13M& 29.14/0.969 & 35.61/0.969 & 33.81/0.874 & 30.84/0.872 & 27.69/0.762 &31.42/0.906&General\\ \cline{2-12}
    &MambaIR~\cite{MambaIR} &ECCV'24&27M& 29.57/0.970 & 35.42/0.969 & 33.88/0.931 & 30.95/0.874 & 27.74/0.793 &31.51/0.907&General\\ \hline \hline
    \multirow{42}{*}{\rotatebox{90}{\textbf{All-in-One}}} 
    &DL\cite{fan2019general} &TPAMI'19&2M&26.92/0.931 & 32.62/0.931 & 33.05/0.914 & 30.41/0.861 & 26.90/0.740 &29.98/0.875&parameterized image operator\\ \cline{2-12}
    &TKMANet~\cite{Unified} &CVPR'22& 29M& 30.41/0.973& 34.94/0.972& 33.02/0.924 &30.31/0.820& 23.80/0.556&30.50/0.849& two-stage knowledge learning\\ \cline{2-12}
    &AirNet~\cite{li2022all}&CVPR'22 &9M&27.94/0.962 &34.90/0.967 &33.92/0.933 &31.26/0.888 &28.00/0.797 &31.20/0.910&contrastive-based\&degradation-guided\\ \cline{2-12}
    &$\text{PIP}_\text{Restormer}$~\cite{li2023prompt}&arXiv'23 &27M& 32.09/0.981 & 38.29/0.984 & 34.24/0.936 & 31.60/0.893 & 28.35/0.806& 32.91/0.920&prompt-in-prompt learning\\ \cline{2-12}
    &IDR~\cite{zhang2023ingredient}&CVPR'23&15M&29.87/0.970&36.03/0.971&33.89/0.931 &31.32/0.884 &28.04/0.798&31.83/0.911&ingredient-oriented learning\\ \cline{2-12}
    &PromptIR~\cite{potlapalli2024promptir}&NeurIPS'23 &36M& 30.58/0.974 & 36.37/0.972 & 33.98/0.933 & 31.31/0.888 & 28.06/0.799&32.06/0.913&prompt for AiOIR\\ \cline{2-12}
    
    &Gridformer~\cite{Wang2023GridFormerRD}&IJCV'23&34M& 30.37/0.970 &37.15/0.972&33.93/0.931 &31.37/0.887& 28.11/0.801&32.19/0.912& transformer with grid structure\\ \cline{2-12}
    &ProRes~\cite{ma2023prores}&arXiv'23&371M&28.38/0.938&33.68/0.954&32.10/0.907& 30.18/0.863& 27.58/0.779&30.38/0.888&degradation-aware visual prompt\\ \cline{2-12}
    &NDR~\cite{Yao2023NeuralDR}&TIP'24&28M&28.64/0.962& 35.42/0.969&34.01/0.932 &31.36/0.887 &28.10/0.798&31.51/0.910&neural degradation representation\\ \cline{2-12}
    % &$\text{Art}_\text{AirNet}$~\cite{wu2024harmony}&ACM MM'24&9M&30.56/0.977&37.74/0.981 &34.02/0.934& 31.37/0.890 &28.12/0.802&32.36/0.917&via multi-task collaboration\\ \cline{2-12}
    &$\text{Art}_\text{PromptIR}$~\cite{wu2024harmony}&ACM MM'24&36M&30.83/0.979 &37.94/0.982 &34.06/0.934& 31.42/0.891 &28.14/0.801&32.49/0.917&via multi-task collaboration\\ \cline{2-12}
    &AnyIR~\cite{Ren2024AnyIR}&arXiv'24&6M& 31.38/0.979&37.90/0.981&33.95/0.933& 31.29/0.889&28.03/0.797&32.51/0.916&local-global gated intertwining\\ \cline{2-12}
    &DaAIR~\cite{Zamfir2024EfficientDA}&arXiv'24&6M&32.30/0.981 &37.10/0.978& 33.92/0.930& 31.26/0.884&28.00/0.792&32.51/0.913 &efficient degradation-aware \\ \cline{2-12}
    % &CAPTNet~\cite{Gao2023PIO} &TCSVT'24&-& 29.28/-&37.86/-&-/-&30.75/-&-/-& -/-&prompt-based\&ingredient-Oriented\\ \cline{2-12}
    
    &MEASNet~\cite{Yu2024MultiExpertAS}&arXiv'24&31M& 31.61/0.981 &39.00/0.985 &34.12/0.935& 31.46/0.892& 28.19/0.803&32.85/0.919&multi-expert adaptive selection\\ \cline{2-12}
    &U-WADN~\cite{Xu2024UnifiedWidthAD}&arXiv'24&6M& 29.21/0.971&35.36/0.968&33.73/0.931& 31.14/0.886 &27.92/0.793&31.47/0.910&unified-width adaptive network\\ \cline{2-12}
    &Shi \emph{et al.}~\cite{Shi2024LearningFD}&arXiv'24&-&29.20/0.972&37.50/0.980&34.59/0.941 &31.83/0.900& 28.46/0.814&32.32/0.921&frequency-aware transformers \\ \cline{2-12}    
    &DyNet~\cite{Dudhane2024DynamicPT}&arXiv'24&16M&31.98/0.981 &38.71/0.983 &34.11/0.936& 31.44/0.892 &28.18/0.803&32.88/0.920&dynamic pre-training\\ \cline{2-12}
    &Hair~\cite{Cao2024HAIRHA}&arXiv'24&29M&30.98/0.979 &38.59/0.983& 34.16/0.935& 31.51/0.892& 28.24/0.803&32.70/0.919&hypernetworks-based\\ \cline{2-12}
    &LoRA-IR~\cite{ai2024taming}&arXiv'24&- & 30.68/0.961& 37.75/0.979 &34.06/0.935& 31.42/0.891& 28.18/0.803& 32.42/0.914&mixture of low-rank experts \\ \cline{2-12}
    &Up-Restorer~\cite{liu2025up} & AAAI'25&28M&30.68/0.977& 36.74/0.978& 33.99/0.933& 31.33/0.888& 28.07/0.799&32.16/0.915&ADMM-solver-based design\\ \cline{2-12}
    % &$\text{TUR}_\text{AirNet}$~\cite{wu2025debiased} & AAAI'25&9M&30.41/0.976& 38.04/0.983 &33.97/0.931 &31.32/0.886 &28.05/0.795 &32.36/0.914& a novel loss function \\ \cline{2-12}
    &$\text{TUR}_\text{PromptIR}$~\cite{wu2025debiased} & AAAI'25&36M&31.17/0.978 &38.57/0.984& 34.06/0.932& 31.40/0.887 &28.13/0.797 &32.67/0.916& a novel loss function \\ \cline{2-12}
    &AdaIR~\cite{cui2024adair} &ICLR'25&29M& 31.06/0.980 & 38.64/0.983 & 34.12/0.935 &31.45/0.892 &28.19/0.802&32.69/0.918&frequency mining and modulation\\ \cline{2-12}
    &DA-RCOT~\cite{tang2024degradation} & TPAMI'25& 50M&31.26/0.977 &38.36/0.983&33.98/0.934&31.33/0.890&28.10/0.801& 32.60/0.917& as an optimal transport problem\\ \cline{2-12}
    &ABAIR~\cite{serrano2024adaptive}& arXiv'24&41M&33.53/0.984& 38.69/0.982 & 34.18/0.935& 31.38/0.890& 28.25/0.804& 33.21/0.919&a three-phase approach\\ \cline{2-12}
    &MoCE-IR~\cite{zamfir2024complexity}&CVPR'25 &25M& 31.34/0.979 &38.57/0.984 &34.11/0.932& 31.45/0.888&28.18/0.800& 32.73/0.917& mixture-of-complexity-experts framework \\ \cline{2-12}
    &MTAIR~\cite{jiang2024multi} & arXiv'24& - & 31.34/0.983&39.15/0.984&34.14/0.936 &31.50/0.893&28.24/0.805&32.87/0.920&Mamba-Transformer cross-dimensional\\ \cline{2-12}
    &$\text{DCPT}_\text{PromptIR}$~\cite{hu2025universal}& ICLR'25&35M&31.91/0.981 &38.43/0.983 &34.17/0.933& 31.53/0.889 &28.30/0.802& 32.87/0.918& learn to classify degradation\\ \cline{2-12}
    &RamIR~\cite{tang2025ramir}& APPL INTELL'25&22M &31.29/0.977&38.16/0.981&34.04/0.931& 31.61/0.891& 28.19/0.801&32.65/0.916&prompt-driven Mamba-based\\ \cline{2-12}
    &Cat-AIR~\cite{jiang2025cat}& arXiv'25&- &31.49/0.980&38.43/0.983& 34.11/0.935 &31.44/0.892 &28.14/0.803& 32.72/0.919&content and task-aware framework\\ \cline{2-12}
    &DSwinIR~\cite{wu2025content}&arXiv'25&24M&31.86/0.980 &37.73/0.983&34.12/0.933& 31.59/0.890 &28.31/0.803&32.72/0.917&deformable sliding window transformer\\ \cline{2-12}
    &$\text{CPL}_\text{PromptIR}$~\cite{gwu2025}&arXiv'25&36M&31.27/0.980& 38.77/0.985&34.15/0.933 &31.50/0.889 &28.23/0.800 &32.78/0.917&contrastive prompt regularization\\ \cline{2-12}
    &AnyIR~\cite{ren2025any} & arXiv'25 & 6M & 31.75/0.981& 38.61/0.984 &34.12/0.936 &31.46/0.893 &28.20/0.804 &32.83/0.920&local-global gated mechanism\\ \cline{2-12}
    &MIRAGE~\cite{ren2025manifold} & arXiv'25 & 10M & 31.86/0.981 &38.94/0.985 &34.12/0.935 &31.46/0.891& 28.19/0.803 &32.91/0.919& within the SPD manifold space\\ \cline{2-12}
    & BaryIR~\cite{tang2025baryir}& arXiv'25& -&31.33/0.980&38.95/0.984&34.16/0.935&31.54/0.892&28.25/0.802&32.85/0.919&in continuous barycenter space\\ \cline{2-12}
    & $\text{SIPL}_\text{PromptIR}$~\cite{wu2025boosti}&arXiv'25&39M& 31.09/0.977 &38.43/0.984&34.12/0.933 &31.48/0.889& 28.22/0.800&32.67/0.917&privilege learning \\ \cline{2-12}
    % &\multicolumn{10}{|c}{\textbf{Methods with the assistance of foundational models}} \\ \cline{2-12}
    &TextPromptIR~\cite{yan2023textual} $\bigstar \bigoplus$ &arXiv'23& 124M+110M& 31.65/0.978 & 38.41/0.982 & 34.17/0.936& 31.52/0.893& 28.26/0.805& 32.80/0.919&textual prompt for AiOIR\\ \cline{2-12}
    &InstructIR-3D~\cite{conde2024high} $\bigstar \bigoplus$&ECCV'24&16M+17M& 30.22/0.959 &37.98/0.978&34.15/0.933&31.52/0.890&28.30/0.804& 32.43/0.913&natural language prompts\\ \cline{2-12}
    % &InstructIR-5D~\cite{conde2024high}&ECCV'24&16M& 27.10/0.956 &36.84/0.973& 34.00/0.931& 31.40/0.887 &28.15/0.798&31.50/0.909&natural language prompts\\ \cline{2-12}
    &$\text{DA-CLIP}_\text{IR-SDE}$~\cite{Luo2023ControllingVM}$\bigstar $&ICLR'24 &49M+125M& 26.83/0.962 &35.85/0.972& 31.43/0.889 &25.05/0.605& 18.33/0.308& 27.50/0.747&degradation-aware vision-language model\\ \cline{2-12}
    &Perceive-IR~\cite{Zhang2024PerceiveIRLT} $\bigstar$&arXiv'24&42M+86M&30.87/0.975 &38.29/0.980&34.13/0.934& 31.53/0.890 &28.31/0.804&32.63/0.917&quality-aware degradation\\ \cline{2-12}
    &VLU-Net~\cite{zeng2025vision}$ \bigstar$& CVPR'25 &35M+88M&30.71/0.980&38.93/0.984 &34.13/0.935& 31.48/0.892&28.23/0.804& 32.70/0.919&vision-language gradient descent-driven\\ \cline{2-12}
    &DFPIR~\cite{tian2025dfpir} $\bigstar \bigoplus$& CVPR'25 & 31M+63M & 31.87/0.980 &38.65/0.982 &34.14/0.935& 31.47/0.893& 28.25/0.806& 32.88/0.919&degradation-aware feature perturbation\\ \cline{2-12}
    % &UniProcessor~\cite{Duan2024UniProcessorAT}&ECCV'24&1002M&31.66/0.979& 38.17/0.982 &34.08/0.935&31.42/0.891 &28.17/0.803&32.70/0.918& support multimodal control\\ 
\Xhline{1.1pt}
\end{tabular}
}

\vspace{-2mm}
\begin{flushleft}
\footnotesize
\textbf{Note:} The dash (``-'') in the ``Params'' column indicates that there is no corresponding source code. $\bigstar$ denotes models assisted by foundation models (e.g., CLIP, BERT, or DINO); $\bigoplus$ denotes models assisted by additional modalities (e.g., text or depth). These notations are consistent across all subsequent tables.
\end{flushleft}
\end{table*}

\subsection{Experiments and Analysis}
To demonstrate the superiority of different AiOIR models, we provide an objective quality comparison in Tab.~\ref{tab:results1}, Tab.~\ref{tab:results2}, Tab.\ref{tab:results3}, Tab.\ref{tab:results4}, Tab.\ref{tab:results5}. The evaluation metrics are composed of PSNR, SSIM~\cite{SSIM}. Specifically, we summarize the experimental results of different methods under five common experimental settings in the field of AiOIR, \ieno, \emph{Setting1} for three-degradation scenarios in Tab.~\ref{tab:results1}: dehazing, deraining, denoising (the noise level $\sigma = 15$, $\sigma = 25$, $\sigma = 50$), \emph{Setting2} for five-degradation scenarios in Tab.~\ref{tab:results2}: dehazing, deraining, denoising (the noise level $\sigma = 25$), deblurring, and low-light enhancement, \emph{Setting3} for synthetic weather-related scenarios (the All-Weather dataset~\cite{valanarasu2022transweather}) in Tab.~\ref{tab:results3}: snow, rain+fog and raindrop, \emph{Setting4} for real-world weather-related scenarios in Tab.~\ref{tab:results4}: haze, rain and snow, and \emph{Setting5} for mixed-degradation scenarios (the CDD-11 dataset~\cite{guo2024onerestore}) in Tab.~\ref{tab:results5}: low-light, haze, rain, snow, low+haze, low+rain, low+snow, haze+rain, haze+snow, low+haze+rain, and low+haze+snow. The results align closely with the original paper.

For \textit{Setting1} in Tab.~\ref{tab:results1}, ABAIR~\cite{serrano2024adaptive} and DFPIR~\cite{tian2025dfpir} achieve the highest average performance both without and with the assistance of foundational models, respectively, indicating strong generalization across diverse degradations. ABAIR attains an average PSNR/SSIM of 32.91/0.920, while DFPIR follows closely with 32.88/0.919, underscoring that careful model design and training strategies are as crucial as leveraging VLMs. For dehazing, ABAIR also achieves the highest PSNR of 33.53, significantly outperforming others, followed by DaAIR~\cite{Zamfir2024EfficientDA} at 32.30 and PIP~\cite{li2023prompt} at 32.09. MTAIR~\cite{jiang2024multi} and MEASNet~\cite{Yu2024MultiExpertAS} stand out with the highest PSNR values of 39.15 and 39.00 for deraining, respectively. Many recent models adopt complex mechanisms such as frequency-aware transformations~\cite{Shi2024LearningFD}, optimal transport~\cite{tang2024degradation}, deep unfolding networks~\cite{zeng2025vision}, and hypernetwork-based architectures~\cite{Cao2024HAIRHA}, suggesting a growing emphasis on specialized designs to tackle varying degradation patterns.

For \textit{Setting2} in Tab.~\ref{tab:results2}, D$^3$Net~\cite{wang2025dynamic} achieves the best average performance across tasks, particularly excelling in dehazing and low-light enhancement. ABAIR and DFPIR also show strong results, with high average scores, indicating their robustness across multiple degradation types. Prompt-based models such as $\text{CPL}_\text{PromptIR}$~\cite{gwu2025} and CyclicPrompt~\cite{liao2025prompt} are among the top performers, suggesting that prompt learning strategies remain a dominant paradigm for tackling multiple degradations simultaneously. Degradation-awareness remains crucial, with models like DaAIR and Perceive-IR~\cite{Zhang2024PerceiveIRLT} tailoring the restoration process to specific degradation characteristics and yielding excellent performance. The trade-off between model size and performance is also evident. Lightweight designs such as TAPE~\cite{liu2022tape} achieve competitive results in certain tasks, whereas more complex architectures like Gridformer~\cite{Wang2023GridFormerRD} deliver more reliable performance across all tasks.

\begin{table*}[htp]
\centering
\caption{Performance Comparisons of AiOIR Models on Five Challenging Datasets. Denoising results are reported for the noise level $\sigma = 25$.}
\label{tab:results2}   
    \renewcommand\arraystretch{1.15}
\resizebox{\textwidth}{!}{
\begin{tabular}{l|lllcccccccl}
    \Xhline{1.1pt}
    \multirow{2}{*}{\textbf{}} &\multirow{2}{*}{\textbf{Method}} & \multirow{2}{*}{\textbf{Venue \& Year}}&\multirow{2}{*}{\textbf{Params}}& \textbf{Dehazing} & \textbf{Deraining} &\textbf{Denoising}&\textbf{Deblurring}&\textbf{Low-light}&\multirow{2}{*}{\textbf{Average}}&\multirow{2}{*}{\textbf{Approach}}\\ &&&&SOTS~\cite{li2019benchmarking} & Rain100L~\cite{yang2017deep-rain100} & BSD68~\cite{BSD100} & Gopro~\cite{Gopro} & LOL~\cite{wei2018deep} &&\\ \Xhline{1.1pt}
    \multirow{5}{*}{\rotatebox{90}{Single}}
    & ADFNet~\cite{ADFNet} &AAAI'23 &8M  & 24.18/0.928 & 32.97/0.943& 31.15/0.882 & 25.79/0.781 & 21.15/0.823 & 27.05/0.871 & Specific \\ \cline{2-12}
    & DehazeFormer~\cite{DehazeFormer} &TIP'23&25M  & 25.31/0.937 & 33.68/0.954 & 30.89/0.880& 25.93/0.785 & 21.31/0.819 & 27.42/0.875 & Specific \\ \cline{2-12}
    & DRSformer~\cite{DRSformer} &CVPR'23& 34M & 24.66/0.931 & 33.45/0.953& 30.97/0.881 & 25.56/0.780 & 21.77/0.821 & 27.28/0.873 & Specific\\ \cline{2-12}
    & HI-Diff~\cite{chen2023hierarchical} &NeurIPS'23& 24M & 25.09/0.935 & 33.26/0.951& 30.61/0.878& 26.48/0.800 & 22.01/0.870 & 27.49/0.887 &Specific \\ \cline{2-12}
    & Retinexformer~\cite{retinexformer} &ICCV'23& 2M & 24.81/0.933 & 32.68/0.940 & 30.84/0.880& 25.09/0.779 & 22.76/0.863 & 27.24/0.873 &Specific  \\ \hline \hline
    \multirow{7}{*}{\rotatebox{90}{Multiple}} 
    & SwinIR~\cite{Liu2021SwinTH} &ICCVW'21&1M  & 21.50/0.891 & 30.78/0.923& 30.59/0.868 & 24.52/0.773 & 17.81/0.723 & 25.04/0.835 & General\\ \cline{2-12}
    & MIRNet-v2~\cite{Zamir2022MIRNetv2} &TPAMI'22&6M &24.03/0.927 & 33.89/0.954 & 30.97/0.881 & 26.30/0.799 & 21.52/0.815 & 27.34/0.875 & General\\ \cline{2-12}
    & DGUNet~\cite{Mou2022DGUNet} &CVPR'22&17M & 24.78/0.940  & 36.62/0.971& 31.10/0.883& 27.25/0.837 & 21.87/0.823& 28.32/0.891 & General\\ \cline{2-12}
    & Restormer~\cite{Zamir2021RestormerET} &CVPR'22& 26M & 24.09/0.927 & 34.81/0.960 & 31.49/0.884&27.22/0.829 & 20.41/0.806 & 27.60/0.881 & General\\ \cline{2-12}
    & NAFNet~\cite{NAFNet} &ECCV'22&17M& 25.23/0.939 & 35.56/0.967&31.02/0.883 &26.53/0.808 &20.49/0.809 &27.76/0.881 & General\\ \cline{2-12}
    & FSNet~\cite{FSNet} &TPAMI'23&13M & 25.53/0.943 & 36.07/0.968& 31.33/0.883 & 28.32/0.869 & 22.29/0.829 & 28.71/0.898& General\\ \cline{2-12}
    & MambaIR~\cite{MambaIR} &ECCV'24&27M & 25.81/0.944 & 36.55/0.971& 31.41/0.884 & 28.61/0.875 & 22.49/0.832 & 28.97/0.901 & General\\ \cline{2-12}  
    \hline \hline
    \multirow{19}{*}{\rotatebox{90}{\textbf{All-in-One}}} 
    &DL~\cite{fan2019general}&TPAMI'19&2M & 20.54/0.826&21.96/0.762 & 23.09/0.745 & 19.86/0.672 & 19.83/0.712 & 21.05/0.743&parameterized image operator\\ \cline{2-12}
    &Transweather~\cite{valanarasu2022transweather}&CVPR'22 &38M& 21.32/0.885&29.43/0.905 & 29.00/0.841 & 25.12/0.757 & 21.21/0.792 & 25.22/0.836& weather type queries\\ \cline{2-12}
    &TAPE~\cite{liu2022tape}&ECCV'22 &1M&  22.16/0.861 &29.67/0.904 & 30.18/0.855 & 24.47/0.763 & 18.97/0.621 & 25.09/0.801 &task-agnostic prior\\ \cline{2-12}
    &AirNet~\cite{li2022all}&CVPR'22&9M & 21.04/0.884 &32.98/0.951 &  30.91/0.882 & 24.35/0.781 & 18.18/0.735 & 25.49/0.846&contrastive-based\&degradation-guided\\ \cline{2-12}
    &IDR~\cite{zhang2023ingredient}&CVPR'23 &15M& 25.24/0.943 &35.63/0.965 &  31.60/0.887 & 27.87/0.846 & 21.34/0.826 & 28.34/0.893 &ingredient-oriented learning\\ \cline{2-12}
    % &$\text{PIP}_\text{NAFNet}$~\cite{li2023prompt}&arXiv'23&18M & 31.75/0.978 & 37.67/0.980 & 31.25/0.878 & 28.08/0.853 & 23.37/0.854& 30.66/0.899 &prompt-in-prompt learning\\ \cline{2-12}
    &$\text{PIP}_\text{Restormer}$~\cite{li2023prompt}&arXiv'23&27M& 32.11/0.979 & 38.09/0.983 & 30.94/0.877 & 28.61/0.861 & 24.06/0.859 & 30.81/0.901&prompt-in-prompt learning\\ \cline{2-12}
    &PromptIR~\cite{potlapalli2024promptir}&NeurIPS'23&36M&26.54/0.949 &36.37/0.970&31.47/0.886&28.71/0.881 &22.68/0.832&29.15/0.904&prompt for AiOIR\\ \cline{2-12}
    &$\text{DASL}_\text{MPRNet}$~\cite{Zhang2023DecompositionAS}&arXiv'23&15M& 25.82/0.947&38.02/0.980&31.57/0.890&26.91/0.823&20.96/0.826& 28.66/0.893&decomposition ascribed synergistic\\ \cline{2-12}
    % &$\text{DASL}_\text{DGUNet}$~\cite{Zhang2023DecompositionAS}&arXiv'23&17M&25.33/0.943&36.96/0.972 &31.23/0.885 &27.23/0.836 &21.78/0.824&28.51/0.892&decomposition ascribed synergistic\\ \cline{2-12}
    % &$\text{DASL}_\text{MPRNet}$~\cite{Zhang2023DecompositionAS}&arXiv'23&5M&23.64/0.924&34.93/0.961&30.99/0.883&26.04/0.788&20.06/0.805&27.13/0.872&decomposition ascribed synergistic\\ \cline{2-12}
    &Gridformer~\cite{Wang2023GridFormerRD}&IJCV'23&34M&26.79/0.951 &36.61/0.971&31.45/0.885 &29.22/0.884 &22.59/0.831 &29.33/0.904& transformer with grid structure\\ \cline{2-12}
    &$\text{Art}_\text{PromptIR}$~\cite{wu2024harmony}&ACM MM'24&36M&29.93/0.908&22.09/0.891& 29.43/0.843 &25.61/0.776 &21.99/0.811& 25.81/0.846&via multi-task collaboration\\ \cline{2-12}
    &DaAIR~\cite{Zamfir2024EfficientDA}&arXiv'24&6M&31.97/0.980 &36.28/0.975&31.07/0.878 &29.51/0.890& 22.38/0.825& 30.24/0.910 &efficient degradation-aware \\ \cline{2-12}
    &AnyIR~\cite{Ren2024AnyIR}&arXiv'24&6M&29.84/0.977&36.91/0.977&31.15/0.882& 26.86/0.822&23.50/0.845&29.65/0.901&local-global gated intertwining\\ \cline{2-12}
    &MEASNet~\cite{Yu2024MultiExpertAS}&arXiv'24&31M&31.05/0.980& 38.32/0.982 &31.40/0.888&29.41/0.890&23.00/0.845& 30.64/0.917&multi-expert adaptive selection\\ \cline{2-12}
    &Hair~\cite{Cao2024HAIRHA}&arXiv'24&29M& 30.62/0.978 &38.11/0.981& 31.49/0.891 &28.52/0.874& 23.12/0.847 &30.37/0.914&hypernetworks-based\\ \cline{2-12}
    &$\text{TUR}_\text{Transweather}$~\cite{wu2025debiased} & AAAI'25&38M&29.68/0.966 &33.09/0.952 &30.40/0.869& 26.63/0.815& 23.02/0.838 &28.56/0.888& a novel loss function \\ \cline{2-12}
    % &$\text{TUR}_\text{AirNet}$~\cite{wu2025debiased} & AAAI'25&9M&27.59/0.954 &33.95/0.962 &30.93/0.875& 26.13/0.801& 17.88/0.772 &27.30/0.873& a novel loss function \\  \cline{2-12}
    &AdaIR~\cite{cui2024adair} &ICLR'25&29M& 30.53/0.978& 38.02/0.981 &31.35/0.889 &28.12/0.858 &23.00/0.845 &30.20/0.910&frequency mining and modulation\\ \cline{2-12}
    &DA-RCOT~\cite{tang2024degradation} & TPAMI'25& 50M&30.96/0.975 &37.87/0.980&31.23/0.888&28.68/0.872&23.25/0.836& 30.40/0.911& as an optimal transport problem\\ \cline{2-12}
    &ABAIR~\cite{serrano2024adaptive}&arXiv'24&41M&33.46/0.983&38.18/0.983&31.38/0.898&29.00/0.878& 24.20/0.865 &31.24/0.921&a three-phase approach\\ \cline{2-12}
    &MoCE-IR~\cite{zamfir2024complexity}&CVPR'25 &25M&30.48/0.974 &38.04/0.982& 31.34/0.887& 30.05/0.899&23.00/0.852& 30.58/0.919& mixture-of-complexity-experts framework \\ \cline{2-12}
    &$\text{DCPT}_\text{PromptIR}$~\cite{hu2025universal}& ICLR'25&35M&30.72/0.977 &37.32/0.978&31.32/0.885&28.84/0.877& 23.35/0.840&30.31/0.911& learn to classify degradation\\ \cline{2-12}
    &RamIR~\cite{tang2025ramir}& APPL INTELL'25& 22M&31.09/0.9787&37.56/0.979&31.44/0.886 &28.82/0.878& 22.02/0.828&30.18/0.910&prompt-driven Mamba-based\\ \cline{2-12}
    &D$^3$Net~\cite{wang2025dynamic}& arXiv'25&38M&32.57/0.965&38.35/0.973&31.73/0.860& 32.70/0.851& 26.49/0.857 &32.36/0.901&cross-domain and dynamic decomposition\\ \cline{2-12}
    &Cat-AIR~\cite{jiang2025cat}& arXiv'25&- & 30.88/0.978&38.21/0.982& 31.38/0.890 &28.91/0.876& 23.46/0.848& 30.57/0.915&content and task-aware framework\\ \cline{2-12}
    &DSwinIR~\cite{wu2025content}&arXiv'25&24M&30.09/0.975 &37.77/0.982&31.34/0.885& 29.17/0.879 &22.64/0.843 &30.19/0.913&deformable sliding window transformer\\ \cline{2-12}
    &$\text{CPL}_\text{PromptIR}$~\cite{gwu2025}&arXiv'25&36M&30.82/0.978 &38.20/0.983&31.41/0.887&28.72/0.874& 23.65/0.855 &30.55/0.915&contrastive prompt regularization\\ \cline{2-12}
    &AnyIR~\cite{ren2025any} & arXiv'25 & 6M & 31.41/0.980& 38.51/0.983 &31.37/0.891& 28.85/0.883 &23.11/0.856& 30.65/0.919&local-global gated mechanism\\ \cline{2-12}
    &MIRAGE~\cite{ren2025manifold} & arXiv'25 & 10M &31.45/0.980 &38.92/0.985& 31.41/0.892 &28.10/0.858& 23.59/0.858& 30.68/0.914&within the SPD manifold space\\ \cline{2-12}
    & BaryIR~\cite{tang2025baryir}& arXiv'25& -&31.12/0.976&38.05/0.981&31.43/0.891&29.30/0.888&23.38/0.852&30.66/0.918&in continuous barycenter space\\ \cline{2-12}
    % &\multicolumn{10}{|c}{\textbf{Methods with the assistance of foundational models}} \\ \cline{2-12}
    &InstructIR-5D~\cite{conde2024high} $\bigstar \bigoplus$ &ECCV'24&16M+17M&  36.84/0.973 &27.10/0.956& 31.40/0.887&29.40/0.886& 23.00/0.836 &29.55/0.907&natural language prompts \\ \cline{2-12}
    &Perceive-IR~\cite{Zhang2024PerceiveIRLT} $\bigstar$ &TIP'25&42M+86M&28.19/0.964 &37.25/0.977&31.44/0.887 &29.46/0.886& 22.88/0.833& 29.84/0.909&quality-aware degradation\\ \cline{2-12}
    &VLU-Net~\cite{zeng2025vision} $\bigstar$& CVPR'25 &35M+88M&30.84/0.980&38.54/0.982&31.43/0.891 &27.46/0.840& 22.29/0.833 &30.11/0.905&vision-language gradient descent-driven\\ \cline{2-12}
    &DFPIR~\cite{tian2025dfpir} $\bigstar \bigoplus$ & CVPR'25 & 31M+63M& 31.64/0.979 &37.62/0.978& 31.29/0.889 &28.82/0.873& 23.82/0.843 &30.64/0.913&degradation-aware feature perturbation\\ \cline{2-12}
    &CyclicPrompt~\cite{liao2025prompt} $\bigstar \bigoplus$ & arXiv'25& 30M+486M&33.02/0.983& 37.03/0.989 &31.33/0.941 &29.42/0.925 &21.79/0.837&30.52/0.935&cyclic prompt-driven universal framework\\ 
    \Xhline{1.1pt}
\end{tabular}}
\end{table*}

\begin{table}[t]
\centering
\caption{Performance Comparisons of AiOIR Models on All-Weather Dataset~\cite{valanarasu2022transweather}. Below the dividing line in the table is AiOIR methods.}
\resizebox{0.48\textwidth}{!}{
  \begin{tabular}{lccccc}
    \Xhline{1.1pt}
    \textbf{Method} & \textbf{Snow} & \textbf{Rain+Fog} & \textbf{Raindrop} & \textbf{Average}&\textbf{Params}\\ \Xhline{1.1pt}
    SwinIR~\cite{Liu2021SwinTH} & 28.18/0.880 & 23.23/0.869 & 30.82/0.904 & 27.41/0.884 & 12M\\  
    MPRNet~\cite{valanarasu2022transweather} & 28.66/0.869 & 30.25/0.914 & 30.99/0.916 & 29.30/0.900 & 16M\\  
    Restormer~\cite{Zamir2021RestormerET} & 29.37/0.881 & 29.22/0.907 & 31.21/0.919 & 29.93/0.902 & 26M\\ \hline  
    All-in-One~\cite{li2020all} & 28.33/0.882 & 24.71/0.898 & 31.12/0.927 & 28.05/0.902&44M\\ 
    Transweather~\cite{valanarasu2022transweather} & 29.31/0.888& 28.83/0.900 & 30.17/0.916 & 29.44/0.901&38M\\  
    AirNet~\cite{li2022all}&27.92/0.858&23.12/0.837 & 28.23/0.892&26.42/0.862&9M \\  
    WGWS-Net~\cite{zhu2023learning} & 28.91/0.856 & 29.28/0.922& 32.01/0.925 & 30.07/0.901&6M\\  
    WeatherDiff~\cite{ozdenizci2023restoring} & 30.09/0.904 & 29.64/0.931& 30.71/0.931& 30.15/0.922&83M\\  
    TKMANet~\cite{Unified} & 30.24/0.902  & 29.92/0.917& 30.99/0.927 & 30.38/0.915&29M\\  
    UtilityIR~\cite{chen2023always} &  29.47/0.879& 31.16/0.927 & 32.01/0.925 & 30.88/0.910&26M\\  
    AWRCP~\cite{ye2023adverse}&31.92/0.934& 31.39/0.933&31.93/0.931&31.75/0.933&-\\ 
    Histoformer~\cite{sun2024restoring}&32.16/0.926&32.08/0.939&33.06/0.944&32.43/0.936 &17M \\
    LoRA-IR~\cite{ai2024taming}& 32.28/0.930& 32.62/0.945& 33.39/0.949& 32.76/0.941&- \\
    $\mathrm{T}^{3}\text{-}\mathrm{DiffWeather}$~\cite{chen2024teaching} $\bigstar \bigoplus$& 32.37/0.936 & 31.99/0.937&32.66/0.941 &32.34/0.938 & 69M+25M\\
    MWFormer~\cite{zhu2024mwformer}&30.92/0.908& 30.27/0.912& 31.91/0.927& 31.03/0.916&170M \\
    $\text{TUR}_\text{Transweather}$~\cite{wu2025debiased}&30.62/0.909&29.75/0.907&31.61/0.933&30.66/0.916&38M\\
    CyclicPrompt~\cite{liao2025prompt} $\bigstar \bigoplus$&32.16/0.927&32.81/0.937 &32.57/0.945&32.51/0.936&30M+486M\\
    DSwinIR~\cite{wu2025content}&32.58/0.931&32.76/0.950&32.88/0.947&32.74/0.943&24M\\
    DA$^2$Diff~\cite{xiong2025da2diff}&31.42/0.916&31.58/0.939&33.01/0.945&31.67/0.933&-\\
    HOGformer~\cite{wu2025beyond}&32.41/0.930&32.89/0.946&32.72/0.945&32.67/0.940&17M\\
    $\text{CPL}_\text{PromptIR}$~\cite{gwu2025}&32.27/0.928&32.16/0.942&32.73/0.943&32.39/0.938&36M\\
    MODEM~\cite{wang2025modemmortonorderdegradationestimation} & 32.52/0.929 & 33.10/0.941& 33.01/0.943& 32.87/0.938&20M\\
    \Xhline{1.1pt}
\end{tabular}}
\label{tab:results3}
\end{table}

For \textit{Setting3} in Tab.~\ref{tab:results3}, MODEM~\cite{wang2025modemmortonorderdegradationestimation} achieves the strongest overall performance by the Morton-order 2D-selective-scan module, with an average PSNR/SSIM of 32.87/0.938. It is closely followed by DSwinIR \cite{wu2025content} and LoRA-IR \cite{ai2024taming}, which deliver averages of 32.74/0.943 and 32.76/0.941, respectively, demonstrating the effectiveness of deformable window attention and mixture-of-experts. By contrast, early methods such as All-in-One \cite{li2020all} yield a much lower average PSNR of 28.05 dB, highlighting the substantial gains achieved by these newer, specialized architectures.

For \textit{Setting4} in Tab.~\ref{tab:results4}, although Transweather~\cite{valanarasu2022transweather} carries the larger parameter count, it struggles to deliver competitive performance. In contrast, though more parameter-efficient, AirNet still falls short in overall restoration quality in the WeatherStream dataset. TKMANet~\cite{Unified} and WGWS-Net~\cite{zhu2023learning} show a notable improvement in both metrics, balancing effectiveness and efficiency. %The best performance is the model proposed by Yang \emph{et al.} \cite{yang2024language}, which leverages language-driven techniques to achieve superior results across all weather conditions. 
Zhu \emph{et al.}~\cite{zhu2023learning} first construct the Real-World Dataset for image restoration under multiple weather conditions of real scenes. Specifically, we test on SPA+~\cite{wang2019SPA-data} for deraining, RealSnow~\cite{zhu2023learning} for desnowing, and REVIDE~\cite{zhang2021learning} for dehazing. DSwinIR achieves superior performance on this dataset. Overall, these results demonstrate a clear progression in multi-weather restoration, suggesting that the field is advancing towards more sophisticated and robust techniques across diverse conditions.

For \textit{Setting5} in Tab.~\ref{tab:results5}, we categorize existing methods into four settings: (a) One-to-One trains on isolated degradations; (b) One-to-Many trains on multiple isolated degradations and tests on them jointly; (c) One-to-Composite trains and tests on both isolated and composite degradations; and (d) Step-by-Step trains on isolated degradations and tests on their combinations step by step. Among the One-to-Composite methods, OneRestore~\cite{guo2024onerestore}, which introduces a novel dataset CDD-11 encompassing both single and composite degradations, achieves a good balance between accuracy and efficiency. In comparison, the AllRestorer model~\cite{mao2024allrestorer} achieves the best performance by adaptive weights and composite scene descriptors. For the Step-by-Step, $\text{CoR}_\text{OneRestore}$ demonstrates the advantage of chain of restoration, even only requiring training on $n$ isolated degradations instead of the full $2^n - 1$ combinations. Overall, models in the one-to-composite and step-by-step categories show a certain degree of generalization and robustness, making them worthy of further exploration.

\begin{table}[t]
\centering
\caption{Performance Comparisons of AiOIR Models on Real-World Weather-Related Datasets.}
\resizebox{0.48\textwidth}{!}{
  \begin{tabular}{lccccc}
    \Xhline{1.1pt}
    \textbf{Method} & \textbf{Haze} & \textbf{Rain} & \textbf{Snow} & \textbf{Average} & \textbf{Params}\\ 
    \Xhline{1.1pt}
    \multicolumn{6}{c}{\textbf{the WeatherStream Dataset~\cite{Zhang2023WeatherStreamLT}}} \\ \hline
    NAFNet~\cite{NAFNet} & 22.20/0.803 & 23.01/0.803 & 22.11/0.826 & 22.44/0.811 & 17M \\  
    GRL~\cite{GRL} & 22.88/0.802 & 23.75/0.805 & 22.59/0.829 & 23.07/0.812 & 3M \\  
    Restormer~\cite{Zamir2021RestormerET} & 22.90/0.803 & 23.67/0.804 & 22.51/0.828 & 22.86/0.812 & 26M \\  
    MPRNet~\cite{MPRNet} & 21.73/0.763 & 21.50/0.791 & 20.74/0.801 & 21.32/0.785 & 16M \\   
    Transweather~\cite{valanarasu2022transweather} & 22.55/0.774 & 22.21/0.772 & 21.79/0.792  & 22.18/0.779 & 38M \\
    AirNet~\cite{li2022all} & 21.56/0.770 & 22.52/0.797 & 21.44/0.812 & 21.84/0.793 & 9M \\ 
    TKMANet~\cite{Unified} & 22.38/0.805 & 23.22/0.795 & 22.25/0.827 & 22.62/0.809 & 29M \\ 
    WGWS-Net~\cite{zhu2023learning} & 22.78/0.800 & 23.80/0.807 & 22.72/0.831 & 23.10/0.813 & 6M \\  
    % LDR~\cite{yang2024language} & 23.11/0.809 & 24.42/0.818 & 23.12/0.838 & 23.55/0.822 & - \\ 
    \hline
    \multicolumn{6}{c}{\textbf{the Real-World Dataset~\cite{zhu2023learning}}} \\ \hline
    TKMANet~\cite{Unified} & 20.10/0.85 & 37.32/0.97 & 29.37/0.88 & 28.93/0.90 & 29M \\
    TransWeather~\cite{valanarasu2022transweather} & 17.33/0.82 & 33.64/0.93 & 26.92/0.86 & 26.71/0.86 & 38M\\
    $\text{TUR}_\text{Transweather}$~\cite{wu2025debiased}& 20.38/0.88 & 39.78/0.98&29.72/0.91&29.96/0.92&38M\\
    WGWS-Net~\cite{zhu2023learning} & 29.46/0.85 & 38.94/0.98 & 33.61/0.93 & 34.01/0.92 & 6M \\
    DSwinIR~\cite{wu2025content} & 30.14/0.89 & 40.60/0.98 & 33.80/0.93 & 34.85/0.93 & 24M \\
    \Xhline{1.1pt}
  \end{tabular}}
\label{tab:results4}
\end{table}

\begin{table}[t]
\centering
\caption{Performance Comparison on CDD-11 Dataset. Methods are categorized into One-to-One, One-to-Many, and One-to-Composite.}
\resizebox{0.48\textwidth}{!}{
  \begin{tabular}{llcccc}
    \Xhline{1.1pt}
    \textbf{Type} & \textbf{Method} & \textbf{Venue \& Year} & \textbf{PSNR} $\uparrow$ & \textbf{SSIM} $\uparrow$ & \textbf{Params} \\
    \Xhline{1.1pt}
    \multirow{8}{*}{One-to-One} 
    & MPRNet~\cite{MPRNet} & CVPR'21 & 25.47 & 0.856 & 16M \\
    & MIRNetv2~\cite{Zamir2022MIRNetv2} & TPAMI'22 & 25.37 & 0.854 & 6M \\
    & Restormer~\cite{Zamir2021RestormerET} & CVPR'22 & 26.99 & 0.865 & 26M \\
    & DGUNet~\cite{Mou2022DGUNet} & CVPR'22 & 25.33 & 0.844 & 18M \\
    & NAFNet~\cite{NAFNet} & ECCV'22 & 26.22 & 0.796 & 17M \\
    & SRUDC~\cite{song2023under} & ICCV'23 &27.64 & 0.860 & 7M \\
    & Fourmer~\cite{zhou2023fourmer} & ICML'23 & 23.44 & 0.789 & 6M \\
    & OKNet~\cite{cui2024omni} & AAAI'24 & 26.33 & 0.861 & 6M \\
    \hline
    \multirow{7}{*}{One-to-Many}
    & AirNet~\cite{li2022all} & CVPR'22 & 23.75 & 0.814 & 9M \\
    & TransWeather~\cite{valanarasu2022transweather} & CVPR'22 & 23.13 & 0.804 & 38M \\
    & WeatherDiff~\cite{ozdenizci2023restoring} & TPAMI'23 & 24.09 & 0.799 & 83M \\
    & PromptIR~\cite{potlapalli2024promptir} & NeurIPS'23 & 25.90 & 0.850 & 36M \\
    & WGWS-Net~\cite{zhu2023learning} & CVPR'23 & 26.96 & 0.861 & 6M \\
    % & InstructIR~\cite{tian2024instruct}&ECCV'24&28.21&0.859&16M\\
    & HAIR~\cite{Cao2024HAIRHA} & arXiv'24 & 27.85 &0.866&29M\\
    \hline
    \multirow{5}{*}{One-to-Composite}
    & OneRestore~\cite{guo2024onerestore} & ECCV'24 & 28.47 & 0.878 & 6M \\
    & AllRestorer~\cite{mao2024allrestorer} $\bigstar \bigoplus$ & arXiv'24 & 33.72 & 0.944 & 12M+88M \\
    & $\text{DA-CLIP}_\text{NAFNet} $ $\bigstar $ &ICLR'24 & 26.56 &0.860& 86M+125M\\
    & MoCE-IR-S~\cite{zamfir2024complexity}&CVPR'25 & 29.05 &0.881 &11M\\
    & $\text{DCPT}_\text{NAFNet}$~\cite{hu2025universal} &ICLR'25 & 28.84 &0.891& 68M\\
    & VL-UR~\cite{liu2025vl}$ \bigstar$ & arXiv'25 & 28.76 & 0.879& -\\
    & MIRAGE~\cite{ren2025manifold} & arXiv'25& 29.33& 0.887&10M\\
    \hline
    \multirow{2}{*}{Step-by-Step}
    & $\text{CoR}_\text{HAIR}$~\cite{cao2024chain} &  arXiv'24 & 28.33 &0.869 & 30M\\
    & $\text{CoR}_\text{OneRestore}$~\cite{cao2024chain} &  arXiv'24 & 28.84& 0.879& 8M\\
    \Xhline{1.1pt}
  \end{tabular}
}
\label{tab:results5}
\end{table}

\section{Challenges and Future Directions}
\label{sec:future}
\subsection{Challenges}
AiOIR models encounter several challenges that limit their effectiveness in real-world applications. Task conflicts arise from differing objectives in denoising, deblurring, and dehazing, complicating simultaneous optimization and leading to inconsistent performance. Additionally, these models struggle with out-of-distribution (OOD) degradations, as real-world images often exhibit a mix of degradation types that do not align with training data. The computational demands of current models hinder deployment on resource-constrained devices, necessitating a balance between restoration quality and efficiency. Furthermore, the reliance on large-scale, high-quality labeled real-world datasets poses challenges due to the resource-intensive nature of data acquisition, resulting in generalization issues. Lastly, most models focus on RGB images, while handling high-dimensional data introduces further complexities, such as increased dimensionality and temporal consistency requirements. Addressing these challenges is essential for improving AiOIR models' practicality and performance.

\subsubsection{Task Conflicts}
In AiOIR, task conflicts arise due to differing objectives across various tasks like denoising, deblurring, and dehazing. Tasks may require opposite optimizations—denoising reduces high-frequency noise, while deblurring enhances high-frequency details. Additionally, data characteristics for different tasks vary, leading to inconsistent performance when training on multiple tasks. 

\subsubsection{OOD Degradations}
AiOIR models face significant difficulty in handling highly diverse and unforeseen image degradations, which can be treated as OOD degradations. In real-world scenarios, images may suffer from a combination of different degradations, such as blur, noise, low resolution, and compression artifacts. At the same time, the degree of each degradation type varies, and may be inconsistent with the sample distribution at training time while testing.

\subsubsection{Model Complexity and Efficiency}
Despite recent advancements in AiOIR, these models are often computationally expensive and complex. Their large size and high computational demands make them impractical for deployment on resource-constrained devices such as mobile phones or embedded systems. Striking a balance between performance and efficiency remains a significant issue, requiring models to maintain high restoration quality without becoming too cumbersome.

\subsubsection{Limited High-Quality Real-world Data}
Many AiOIR models depend on large-scale, high-quality labeled datasets for supervised training, but acquiring these datasets is resource-intensive. Real-world degraded data is often scarce, and the unpredictability of degradations makes it difficult for models to perform well in practical applications. 

\subsubsection{High Dimensional Data}
Current IR models primarily focus on 2D images, but handling 3D data and video sequences presents additional challenges. For video restoration, not only does each frame need high-quality restoration, but temporal consistency across frames must also be preserved. This adds complexity and demands more sophisticated methods that can integrate spatial and temporal information simultaneously.

\subsection{Future Research Directions}

Future research will pursue several directions: developing a robust multi-task learning theory to resolve task conflicts and enhance information sharing; exploring semi‑ and unsupervised methods to reduce dependence on large labeled datasets; designing efficient, edge‑compatible models for practical deployment; examining complex real‑world degradation scenarios to boost performance; and integrating large multimodal pretrained models with generative priors. Together, these efforts will underpin more practical, flexible AiOIR models capable of addressing diverse real‑world challenges.

\subsubsection{Incorporating Robust Multi-task Learning Theory} The development of multi-task learning (MTL) theory for AiOIR is still in its early stages, presenting key research opportunities \cite{zhang2023ingredient, zhang2024real}. Challenges include managing task conflicts, dynamically weighting tasks, and optimizing inter-task information sharing to avoid interference \cite{Yu2020GradientSF,Liu2021ConflictAverseGD}. Further improvements can come from determining optimal task sequences, prioritizing tasks based on degradation severity, and designing loss functions that balance conflicting objectives. Progress in these areas is essential for building efficient, generalizable AiOIR systems for real-world degradations.

\subsubsection{Robust Prompt Learning under Data Scarcity} While prompt-based learning enhances adaptability in AiOIR, its tendency to overfit under limited or imbalanced training data is a key challenge. Future work may explore prompt-space regularization (e.g., sparsity or low-rank constraints) to limit over-specialization, as well as meta-learning frameworks (e.g., MAML) to promote generalization across unseen degradations. Incorporating degradation-aware conditioning can further improve robustness. Overall, designing prompt mechanisms that generalize reliably remains an open and important direction.

\subsubsection{Semi-Supervised and Unsupervised Learning Methods} Reducing the dependency on large-scale labeled datasets is essential for the scalability and applicability of AiOIR models. Future research should focus on developing semi-supervised and unsupervised learning approaches that can learn effective representations from unlabeled or partially labeled data \cite{poirier2023robust,xie2024learning}. Techniques such as self-supervised learning, contrastive learning, and unsupervised domain adaptation can be leveraged to improve model performance in scenarios where labeled data is scarce or unavailable. By advancing these methods, AiOIR models can become more adaptable to diverse and unforeseen degradations encountered in real-world conditions.

\subsubsection{Platform-Aware Model Design and Efficient Approaches} An important direction is developing edge-friendly AiOIR models that balance accuracy and efficiency for deployment on mobile and embedded devices \cite{michelini2022edge, zhang2021edge}. Techniques like compression, pruning, quantization, and architecture search help reduce resource use with minimal performance loss. In addition to these methods, more radical approaches such as binary neural networks (BNNs) and spiking neural networks (SNNs) offer promising avenues for deployment. Recent works show their effectiveness in tasks like super-resolution and deraining \cite{xia2022basic, wei2023ebsr, Song2024LearningAS, Su2025BridgeTG}. Combined with lightweight transformers, edge-specific distillation, and early-exit strategies, these methods advance practical AiOIR deployment.

\subsubsection{Addressing More Practical and Complex Degradations} There is a need to focus on more practical tasks and datasets that reflect the complexities of real-world image degradations \cite{li2023lsdir,yu2024scaling}. Future research should explore composite and complex degradation scenarios, such as image restoration in nighttime conditions, deraining and dehazing in the dark, and images affected by multiple overlapping degradations rather than isolated, mixed single degradation tasks. Developing and utilizing datasets that capture these challenging conditions will enable models to learn from and be tested on data that closely resembles real-world challenges. This focus will drive the development of AiOIR models that are more robust and effective in practical applications.

\subsubsection{Incorporating Large Multimodal Pretrained Models and Generative Priors} Another promising avenue is exploiting large multimodal pretrained models, particularly those incorporating generative models, to enhance AiOIR tasks \cite{wu2024q, jin2024llmra, xu2024boosting, zheng2024lm4lv}. Models like CLIP~\cite{radford2021learning} and recent advancements in generative models (e.g., stable diffusion \cite{rombach2021highresolution}) have demonstrated remarkable abilities to capture complex data distributions across multiple modalities. By mining the rich representations and priors from these universal models, AiOIR can benefit from enhanced understanding of image content and context, leading to better restoration in low-level tasks.

\subsubsection{Leveraging Multi-Modal Information} Most current AiOIR models primarily rely on single-modal image information, such as RGB images, limiting their effectiveness in addressing complex restoration tasks. Future research can focus on integrating multi-modal information—like depth maps, optical flow, and infrared images—into AiOIR frameworks \cite{deng2020deep,zhang2024unified,zhao2024equivariant}. This integration would provide models with rich contextual and structural insights, enhancing their ability to accurately recover images with varied degradations \cite{ma2019infrared}. 

\subsubsection{Establishment of Standardized Evaluation Protocols and Benchmarks} Unlike single image restoration tasks with established datasets, AiOIR lacks standardized benchmarks. To enable fair comparison, it's essential to develop unified evaluation protocols and diverse datasets covering real-world scenarios like high-resolution restoration, medical enhancement, old photo repair, and challenging weather conditions.

\subsubsection{Extension to High-dimensional Data} In addition to RGB images, extending AiOIR methods to other data, such as video \cite{liang2024vrt, zhou2023nerflix}, 3D data \cite{zhou2023nerflix,li2024hierarchical}, dynamic event data \cite{wang2020eventsr,liang2023event}, and hyperspectral data \cite{pang2024hir}, presents significant opportunities for future research. Developing techniques that effectively integrate spatial, temporal, spectral, dynamic, and 3D information will be vital for applications such as video enhancement, 3D rendering, spectral analysis, and augmented reality. Addressing these challenges will expand the capabilities of AiOIR models, making them more versatile and applicable to a broader range of tasks.

\subsubsection{Privacy Protection} AiOIR is increasingly deployed in medical imaging, surveillance, and autonomous driving, yet its ethical implications are rarely addressed. Prior work on machine unlearning~\cite{Su2024AccurateFF} shows how to erase the influence of private data in AiOIR, suggesting future research can explore data deletion and privacy-preserving training. Beyond privacy, biased data or model priors can cause fairness issues like uneven performance or hallucinations. Furthermore, misuse risks must be considered, as restored content may be used maliciously in misinformation, surveillance overreach, or unauthorized identity synthesis. These issues highlight the need for responsible development and deployment.

\section{Conclusion}
\label{sec:conclusion}
In this paper, we present a comprehensive review of recent advancements in all-in-one image restoration (AiOIR), a rapidly emerging domain that integrates multiple types of image degradations into a single framework. Through an in-depth exploration of state-of-the-art models, we emphasize their robust capabilities, diverse architectures, and the rigorous experiments. By contrasting these models with traditional single-task approaches, we underscore the limitations of the latter in addressing real-world complexities, while highlighting the significant gains in efficiency, adaptability, and scalability offered by AiOIR models.

Our proposed taxonomy offers a multi-faceted perspective on this rapidly evolving field, covering network innovations, key methodologies such as prompt learning, mixture-of-experts, the incorporation of multimodal models, and autonomous agents. We further present an analysis of critical datasets, providing researchers and practitioners with a holistic toolkit to better assess the current landscape of AiOIR. Despite notable advancements, persistent challenges remain: current models still struggle with handling complex and compounded degradations, suffer from computational inefficiencies, and do not generalize well in real-world scenarios. Looking ahead, we anticipate that the trajectory of AiOIR research will be shaped by several converging trends: the development of more lightweight and efficient architectures, advancements in semi-supervised learning, and expanding the scope of AiOIR models to accommodate multimodal inputs and video data. Furthermore, breakthroughs in real-time processing, cross-modal reasoning, and model interpretability are poised to play a pivotal role in unlocking the full potential of unified restoration frameworks.

In conclusion, AiOIR is not merely a technical amalgamation of multiple restoration tasks—it represents a foundational rethinking of how visual systems can perceive, process, and reconstruct information in complex environments. By charting the contours of this evolving landscape, our review seeks not only to synthesize current knowledge but also to catalyze future innovations at the intersection of efficient computation, multimodal understanding, and autonomous adaptation. As AiOIR continues to mature, it holds the promise of becoming a cornerstone technology for general-purpose visual intelligence in the wild.

\bibliographystyle{IEEEtran}
\bibliography{refs}
\vspace{-13 mm} 
\begin{IEEEbiography}[{\includegraphics[width=1.0in,height=1.25in,clip,keepaspectratio]{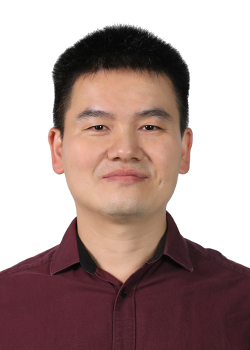}}]{Junjun Jiang}
received the B.S. degree in Mathematics from the Huaqiao University, Quanzhou, China, in 2009, and the Ph.D. degree in Computer Science from the Wuhan University, Wuhan, China, in 2014. 
He is currently a Professor with the School of Computer Science and Technology, Harbin Institute of Technology, Harbin, China. His research interests include image processing and computer vision. 
\end{IEEEbiography}
\vspace{-13 mm} 

\begin{IEEEbiography}[{\includegraphics[width=1in,height=1.25in,clip,keepaspectratio]{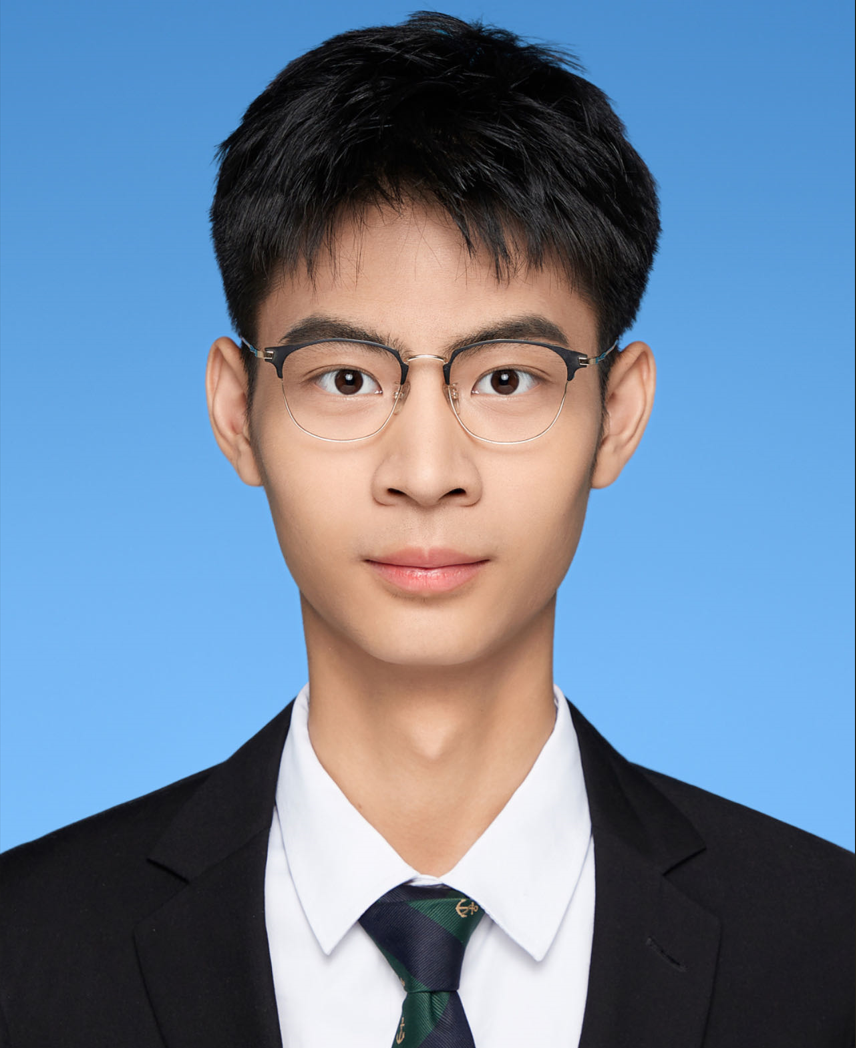}}]{Zengyuan Zuo} received the B.E. degree in the Harbin Institute of Technology (HIT), Harbin, China, in 2024. He is currently pursuing the MA.Eng. degree in Faculty of Computing at Harbin Institute of Technology. His research interests include image restoration. 
\end{IEEEbiography}
\vspace{-13 mm} 

\begin{IEEEbiography}[{\includegraphics[width=1.0in,height=1.25in,clip,keepaspectratio]{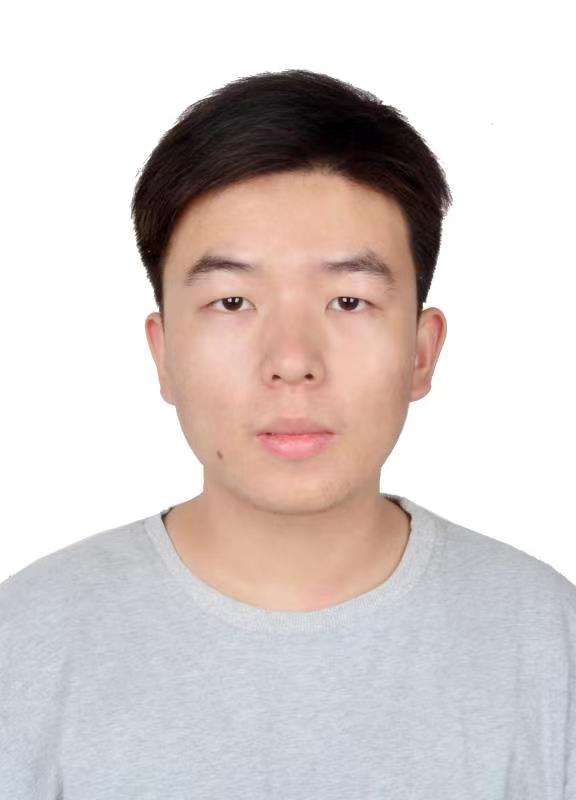}}]{Gang Wu} received the B.E. degree in the School of Computer Science and Technology from Soochow University, Jiangsu, China, in 2020. He is currently pursuing the Ph.D. degree in Faculty of Computing at Harbin Institute of Technology. His research interests include image restoration, representation learning, and self-supervised learning.
\end{IEEEbiography}
\vspace{-13 mm} 

\begin{IEEEbiography}[{\includegraphics[width=1in,height=1.25in,clip,keepaspectratio]{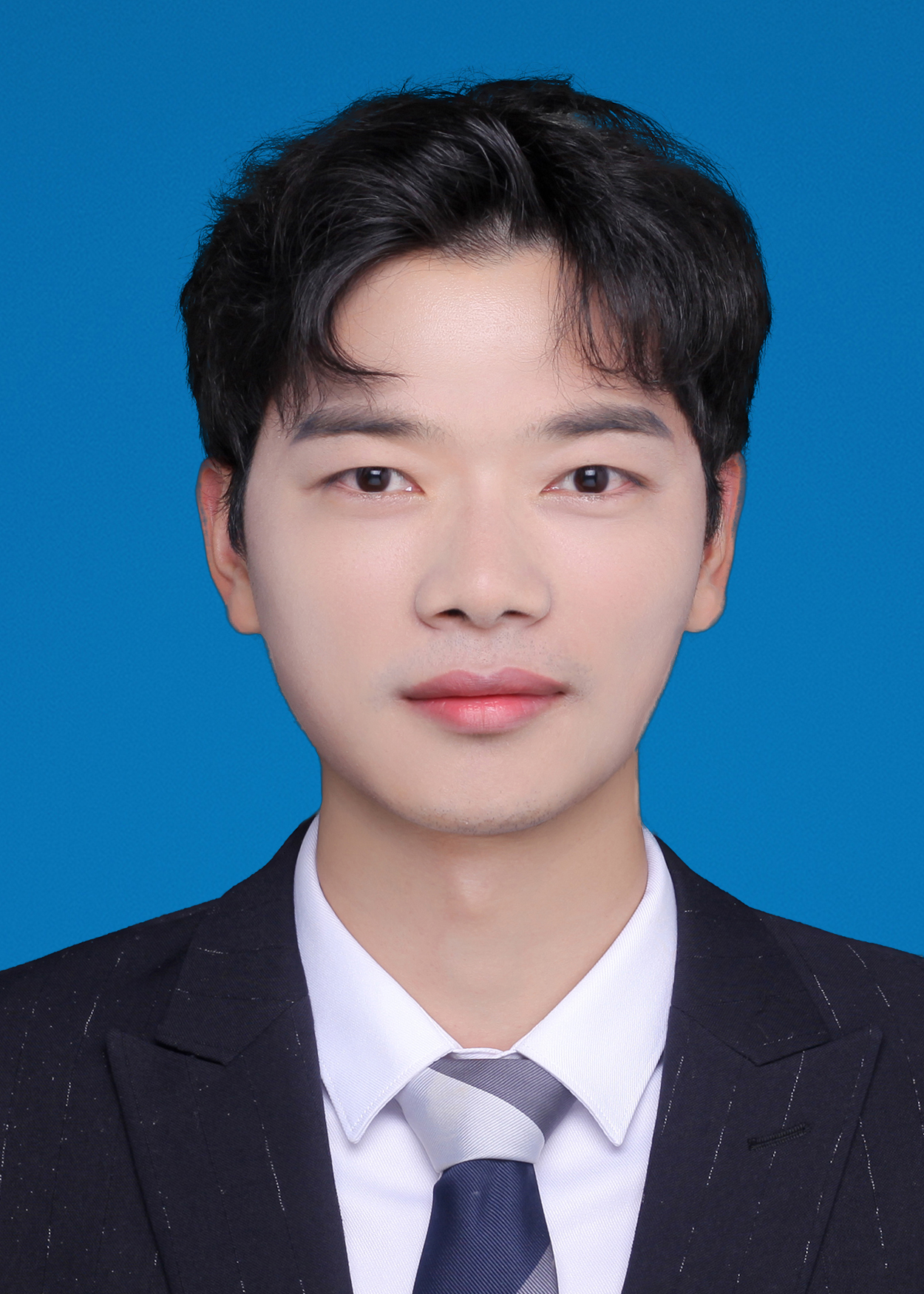}}]{Kui Jiang} received the Ph.D. degree in the school of Computer Science, Wuhan University, Wuhan, China, in 2022. He is currently an associate professor with the school of Computer Science and Technology, Harbin Institute of Technology, Harbin, China. His research interests include image/video processing and computer vision. 
\end{IEEEbiography}
\vspace{-13 mm} 

\begin{IEEEbiography}[{\includegraphics[width=1.0in,height=1.25in,clip,keepaspectratio]{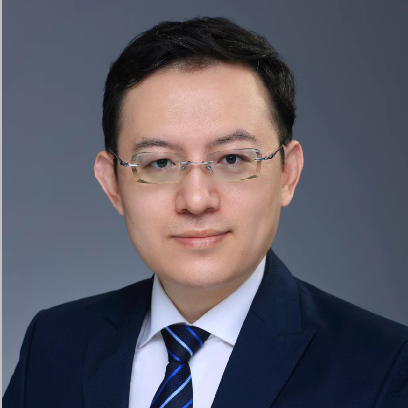}}]{Xianming Liu}
received the B.S., M.S., and Ph.D. degrees in computer science from the Harbin
Institute of Technology (HIT), Harbin, China,
in 2006, 2008, and 2012, respectively. He is currently a Professor with the School of Computer Science and Technology, HIT. His research interests include trustworthy AI, 3D signal processing and biomedical signal processing.
\end{IEEEbiography}

\end{document}